\documentclass[sigconf]{acmart}

\AtBeginDocument{%
  \providecommand\BibTeX{{%
    \normalfont B\kern-0.5em{\scshape i\kern-0.25em b}\kern-0.8em\TeX}}}

\copyrightyear{2020}
\acmYear{2020}
\setcopyright{acmlicensed}
\acmConference[MM '20]{Proceedings of the 28th ACM International Conference on Multimedia}{October 12--16, 2020}{Seattle, WA, USA}
\acmBooktitle{Proceedings of the 28th ACM International Conference on Multimedia (MM '20), October 12--16, 2020, Seattle, WA, USA}
\acmPrice{15.00}
\acmDOI{10.1145/3394171.3413972}
\acmISBN{978-1-4503-7988-5/20/10}
\usepackage{float}
\usepackage{multirow}
\usepackage{soul}
\usepackage[utf8]{inputenc}
\usepackage{caption}
\usepackage{subcaption}
\usepackage{graphicx}
\usepackage{mwe}
\usepackage{array}
\usepackage{booktabs,caption}
\usepackage{threeparttable}
\usepackage{float}
\usepackage{bm}
\usepackage{multirow}
\usepackage{booktabs}
\usepackage{algorithm}
\usepackage{algorithmic}
\usepackage{amsmath}
\usepackage{amsthm}

\newtheorem{proposition}{Proposition}
\newcommand{\tabincell}[2]{\begin{tabular}{@{}#1@{}}#2\end{tabular}}
\usepackage{url}

\begin{document}
\fancyhead{}
\title{Privacy-sensitive Objects Pixelation for Live Video Streaming}

\author{Jizhe Zhou}
\affiliation{%
  \institution{Department of Computer and Information Science, University of Macau}
  \streetaddress{Avenida da Universidade,Taipa}
  \city{Macau}
  \state{China}
  \postcode{999078}
}
\email{yb87409@um.edu.mo}

\author{Chi-Man Pun}
\authornote{Chi-Man Pun is the corresponding author}
\affiliation{%
    \institution{Department of Computer and Information Science, University of Macau}
  \streetaddress{Avenida da Universidade,Taipa}
  \city{Macau}
  \state{China}
  }
\email{cmpun@umac.mo}

\author{Yu Tong}
\affiliation{%
  \institution{vivo AI Lab, Shenzhen \& Department of Computer and Information Science, University of Macau}
  \streetaddress{Avenida da Universidade,Taipa}
  \city{Macau}
  \state{China}
  }
\email{yb87462@um.edu.mo}
\begin{abstract}
With the prevailing of live video streaming, establishing an online pixelation method for privacy-sensitive objects is an urgency. Caused by the inaccurate detection of privacy-sensitive objects, simply migrating the tracking-by-detection structure applied in offline pixelation into the online form will incur problems in target initialization, drifting, and over-pixelation. To cope with the inevitable but impacting detection issue, we propose a novel Privacy-sensitive Objects Pixelation (PsOP) framework for automatic personal privacy filtering during live video streaming. Leveraging pre-trained detection networks as the backbone, our PsOP is extendable to any potential privacy-sensitive objects pixelation. Employing the embedding networks and the proposed Positioned Incremental Affinity Propagation (PIAP) clustering algorithm, our PsOP unifies the pixelation of discriminate and indiscriminate pixelation objects through trajectories generation. In addition to the pixelation accuracy boosting, experiment results on the streaming video data we built show that the proposed PsOP can significantly reduce the over-pixelation ratio and the human intervention in privacy-sensitive object pixelation.
\end{abstract}

\begin{CCSXML}
<ccs2012>
	<concept>
		<concept_id>10002978.10003029.10011150</concept_id>
		<concept_desc>Security and privacy~Privacy protections</concept_desc>
		<concept_significance>500</concept_significance>
	</concept>
	<concept>
		<concept_id>10003752.10010070.10010071.10010074</concept_id>
		<concept_desc>Theory of computation~Unsupervised learning and clustering</concept_desc>
		<concept_significance>500</concept_significance>
	</concept>
	<concept>
		<concept_id>10010147.10010178.10010224.10010245.10010253</concept_id>
		<concept_desc>Computing methodologies~Tracking</concept_desc>
		<concept_significance>500</concept_significance>
	</concept>
	<concept>
		<concept_id>10010147.10010178.10010224.10010225.10010227</concept_id>
		<concept_desc>Computing methodologies~Scene understanding</concept_desc>
		<concept_significance>300</concept_significance>
	</concept>
</ccs2012>
\end{CCSXML}

\ccsdesc[500]{Security and privacy~Privacy protections}
\ccsdesc[500]{Theory of computation~Unsupervised learning and clustering}
\ccsdesc[500]{Computing methodologies~Tracking}
\ccsdesc[300]{Computing methodologies~Scene understanding}
\keywords{object pixelation; neural networks; PIAP clustering; live video-streaming}
\maketitle

\section{Introduction}

\begin{figure}[htbp]
\centering
\begin{subfigure}[b]{0.5\textwidth}
		\includegraphics[scale=0.22]{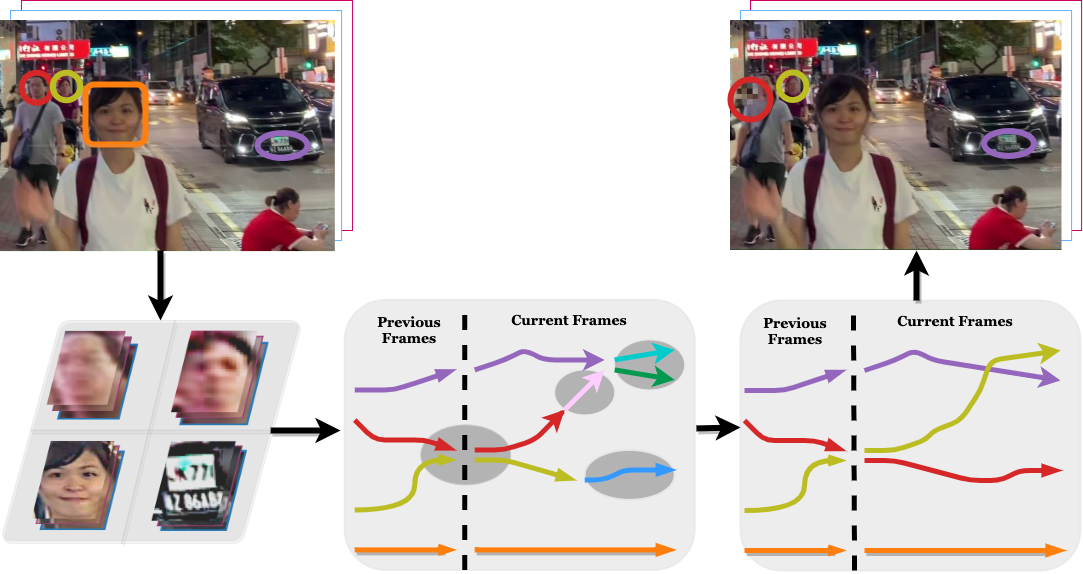}
		\caption{Established offline pixelation pipeline.}
		\vspace{0.5em}
        %\label{fig:y equals x}
\end{subfigure}
\begin{subfigure}[b]{0.5\textwidth}
		\includegraphics[scale=0.22]{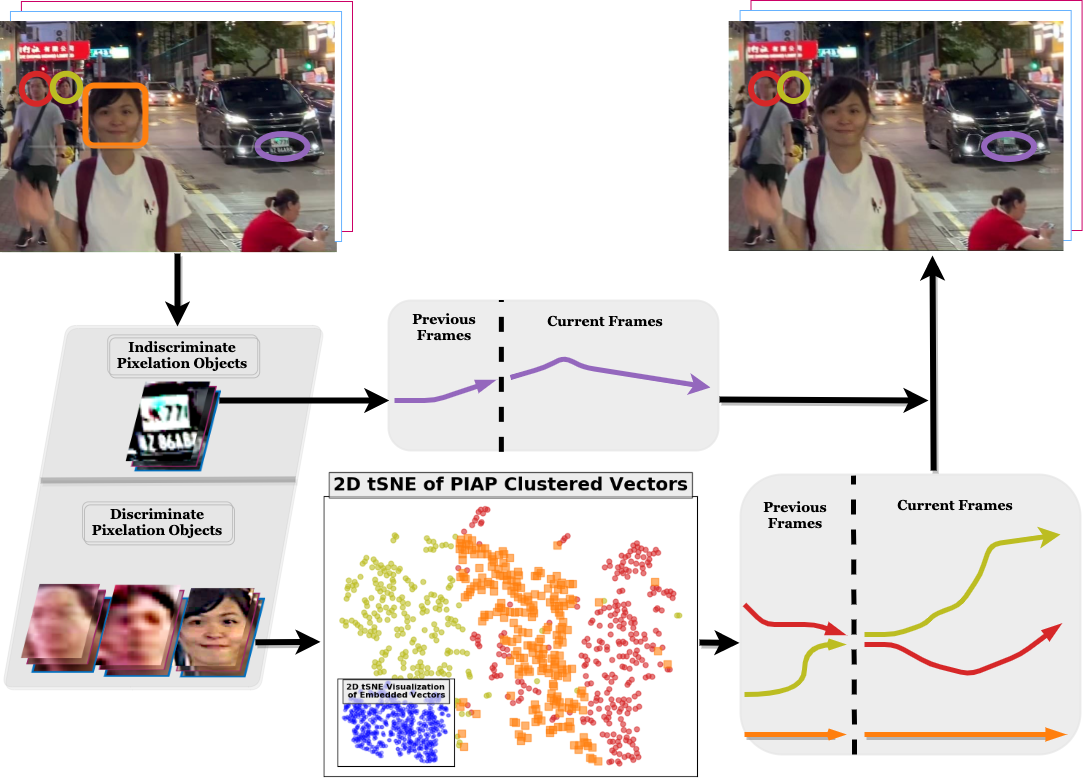}
		\caption{Established offline pixelation pipeline.}
		\vspace{0.5em}
        %\label{fig:y equals x}
\end{subfigure}
\caption{Current pixelation methods v.s. PsOP under the same scene. The leftmost clips are a clip of a streaming scene. Potential privacy-sensitive objects are labeled in circles and rectangular and only circled objects shall be pixelated. Tracklets rendered in gray ground require manual association in (a). The color of the clusters in (b) represents the corresponding colored objects. Pixelation results are shown in the rightmost clip.}
\end{figure}
\iffalse
	\subfloat[Established tracking-by-detection pixelation pipeline]{
		\includegraphics[scale=0.22]{img/testa.png}}
		%\caption{Established offline pixelation pipeline.}
		\vspace{0.5em}
        %\label{fig:y equals x}
	\subfloat[PsOP pixelation process]{
		\includegraphics[scale=0.22]{img/testb.png}}
		%\caption{Propose PsOP working process.}
        %\label{fig:y equals x}
	%\end{subfigure}
%\includegraphics[scale=0.22]{img/test2.png}
\fi
Live video streaming has never been so popular as in the past two or three years. Live video streaming, especially the outdoors, presents audiences unprepared, unrehearsed filming of reality, and ramps up the sense of immersion, unpredictability or nervousness. Real scenes are recorded, then instantly broadcast to audiences mostly through streamers' phone cameras and the mobile network. Streaming activities join a symbiosis with the popularization of smartphones, 5G networks, and cheaper communication costs~\cite{stewart2016up}.  The primary host of live streaming activities transformed from TV stations and news bureau to ordinary individuals. Every rose has its thorn. Without imposed censorship or filters before broadcasting, the pervasiveness of live streaming today arouses unprecedented violations in the personal privacy protection field. Strict law or conservative religions even commit those violations to crimes~\cite{faklaris2016legal}.

\par
Actually, apart from the indifference of streamers and the absence of imposed censorship, the handcrafted pixelation process is the main reason that leads to privacy infringements amid video streaming. As a giant live-streaming hosting service provider, YouTube is already aware of such urgent needs and published its offline auto-pixelation tool on the latest YouTube Creator Studio. As far as we know, current studies, including YouTube Studio and Microsoft Azure, mainly focus on processing offline videos; whilst leave the pixelation methods on the online live video streaming field underexplored. To ensure the privacy rights and the sound development of the streaming industry, proper pixelation methods shall be interpolated between the filming and broadcasting of live video streaming. Transparency in terms of user experience is also indispensable to keep the pixelation practical. Therefore, in this paper, we devote to establish a pixelation method that generates automatic personal privacy filtering during unconstrained streaming activities.

\par
Most of the few existing offline pixelation methods adopt a similar tracking-by-detection structure that assembles privacy-sensitive objects detectors ahead of multi-object tackers. In such a structure, a Multi-Object Tracker (MOT) aims at continuously locating an arbitrary target in a video with a given bounding box in the target's initial frame, while the detector is responsible for giving the bounding box in the target's initial frame. As shown in Figure 1 (a), this structure intuitively follows the manual pixelation pipeline and works well on fine-shot videos. Once a sensitive object is spotted by the detector, it is tracked by the tracker till its vanishment. This action sequence loops over for every detection. Allocating Gaussian or other filters on the trajectories reaches the final pixelation. Essentially, the object trajectories determine the final pixelation results.

\par
In tracking-by-detection structure, trajectories are the joint effort of trackers and detectors, and in fact, trackers and detectors are contained independently. Comparing with conventional fine-shot videos, live streaming videos always involve very few shot changes and the shaky camera. Recorded often by hand-held cameras, shaky or jerky camera stands for frequent camera shakes, abrupt streaming condition alteration, and noisy backgrounds. Under such conditions, trackers still struggle standing up to scattered or drifted tracklets in the long-term. Even were the long-term tracking accuracy somehow resolved, tracking-by-detection structure also prone to corruption as its performance decisively in accordance with detection accuracy. Trackers are unable to function well with poor initiations. However, due to inadequate training samples and the lack of comprehension for video contexts, false positives and false negatives produced by the image-based detectors spring up in streaming videos. Consequently, besides efficiency, blinking, drifting, missing, or excessive mosaics are allocated everywhere when migrating existing methods to pixelate live videos. Hence, blaming unacceptable mosaics mainly on inaccurate detections, current methods are incapable of handling live video streaming.
%Scattered and drifted tacklets lead to blinks and drifts of mosaics unless an intensive manual revision is involved.
%Hence, stationary images are inadequate in training video-level privacy-sensitive objects detectors. The widespread false negatives and positives in detection instantly cause the mosaics blinks.mosaics yielded by current methods always contains frequent blinks, drifts,  Generally, the final trajectories ready for pixelation always contain numerous interventions and unfixable drifting. Thus, besides the efficiency, frequent interventions or drifting make these methods unrealistic in dealing with live video streaming.
%Existing pixelation methods can be described as an aggregation of parallel single-object trackers built on the tracking-by-detection (of sensitive objects) algorithm,

%Consequently, pixelating on trajectories initialized by inaccurate detection escalates the yielded mosaics' blinking and drifting. Generally, the final trajectories ready for pixelation always contain numerous interventions and drifting. Thus, besides the efficiency, frequent interventions or drifting make these methods unrealistic in dealing with live video streaming.
%the accuracies of these detectors suffer drastic drops and tend to overestimate the areas needing pixelation.
\par
The other major drawback of the current pixelation workflow is the over-pixelation problem. Introduced by the tracking algorithms and inherited by current tracking-by-detection pixelation methods, the over-pixelation problem occurs when the trackers insist on generating puzzling and unnecessary mosaics for unidentifiable privacy-sensitive objects. Heavy or fully occlusion, and massive motion blur are the common reasons for objects' unidentifiable. Tracking algorithms are designed to construct seamless or long enough trajectories for objects in avoiding frequent ID-switch. Unlike tracking, in the meantime of blocking the privacy-sensitive information from leaking, pixelation tasks are dedicated to preserving the audience as many originals as possible. Over-pixelation is an intrinsic problem while migrating tracking algorithms to pixelation tasks.
\par
To address the aforementioned issues, our Privacy-sensitive Objects Pixelation (PsOP) introduces a brand-new framework for generating reliable pixelation with real-time efficiency in live video streaming. To yield reliable trajectories for pixelation, PsOP foremost copes with the inaccurate detections caused by deep networks insufficiency and the lack of comprehension for video contexts. PsOP divides the potential privacy-sensitive objects into contexts-irrelevant Indiscriminating Pixelation Objects (IPOs) and contexts-relevant Discriminating Pixelation Objects (DPOs) to alleviate contexts' effects on detection. As their sensitiveness barely changes in scenes, IPOs, including erotic images, trademarks, phone numbers or car plate numbers, etc., are liable to be handled through well-established sensitive objects detection algorithms~\cite{yu2016iprivacy}. Conversely, DPOs mainly made up of faces and texts are not to be solely dealt with detection networks. When the domain of discriminating and indiscriminating pixelation objects overlaps (e.g., the phone number is also a kind of text), IPOs are prioritized in claiming the overlapped bounding box for tighter privacy protection. Depicted in Figure 1 (b), detected IPOs are smoothed and then pixelated.

\par
The remaining Discriminating Pixelation Objects (DPOs) are the primary concern of PsOP. The sensitiveness of instances belonging to DPOs changes according to scenes or contexts. The varying sensitiveness of DPOs along with the inbuilt network insufficiency cannot be simply solved through training on video data. Such training assumes the detection, recognition, and video semantics segmentation tasks share a unified network structure as well as demands labor-intensive video data labeling. Considering the learning-based feature vectors are in fact the central role for instances' trajectories association, as shown in Figure 1 (b), PsOP abnegates tracking-by-detection structure and employs the detection, embedding, and clustering procedure to generate trajectories under inaccurate detections.
\par
In specific, for every class of DPOs, pre-trained detection and embedding networks are leveraged to yield feature vectors for instances of the class. Subsequently, we propose the Positioned Incremental Affinity Propagation (PIAP) algorithm for clustering under inaccurate detection and embeddings. In classic Affinity Propagation (AP), affinities are the distances between feature vectors. A series of messages about affinities, denoted as availabilities and responsibilities, are propagated among vectors to reach a final consensus on clustering results. For PsOP, since an individual instance of DPOs cannot appear at different positions within a frame, the detected instances positioned in the same frame serve as negative samples with minimal affinities. These minimized affinities revise the consensus through the propagation of availabilities and responsibilities. The preference matrix is also computed based on availabilities and responsibilities. Weak preference indicates the detection is relatively isolated in the feature space; thereby, excluded as outliers through propagation. Besides, as a single instance owns similar expanding availability and responsibility matrices in adjacent frames, positioned affinities are propagated in an incremental way. A run of PIAP iterates until the consensus of clusters' results is reached for all vectors. Link the detection within the same cluster forms the trajectory of an instance. Comparing with classic AP, while retaining the ability of clustering under ill-defined cluster number, PIAP solves the noise-sensitive and time-consuming problem of AP by introducing positioned affinities and incremental propagation. Moreover, unlike gradually fitting trackers, detection, and embedding based PIAP simultaneously solves the inherent over-pixelation problem.

\par
In order to evaluate the performance of the proposed PsOP and PIAP, we crawled raw streaming videos from YouTube Live and Facebook Live platforms. These raw videos contents are inspected and selected in detail to ensure the dataset diversities. Tested on the live streaming video dataset (84261 labeled boxes and 19372 frames) we collected and manually labeled, PsOP gains significant pixelation accuracy boosting as well as retains much more non-sensitive originals for audiences.

\par
In summary, the main contributions of this paper are as follows:
\begin{itemize}%\vspace{-0.43em}
    \item We build the Privacy-sensitive Objects Pixelation (PsOP) framework for the pixelation of privacy-sensitive objects in live video streaming. As far as we know, PsOP is the first online method adopts the detection, embedding, clustering procedures and solves the over-pixelation problem.
    \item We proposed the Positioned Incremental Affinity Propagation (PIAP) clustering algorithm to generate trajectories for inaccurately detected and sensitiveness varying discriminating pixelation objects. The proposed PIAP spontaneously handles the cluster number generation, cluster under unbalanced sample-size problems, and further endows the classic AP with noise-resistance and time-saving merits.
    \item We built a live video streaming dataset to test the proposed PsOP framework and to be available to public. Diverse streaming videos are collected through live streaming platforms, and dense annotations on each frame are manually labeled for the evaluation of the proposed method.
    \end{itemize}
\par

\begin{figure*}[htbp]
\centering
\includegraphics[scale=0.27]{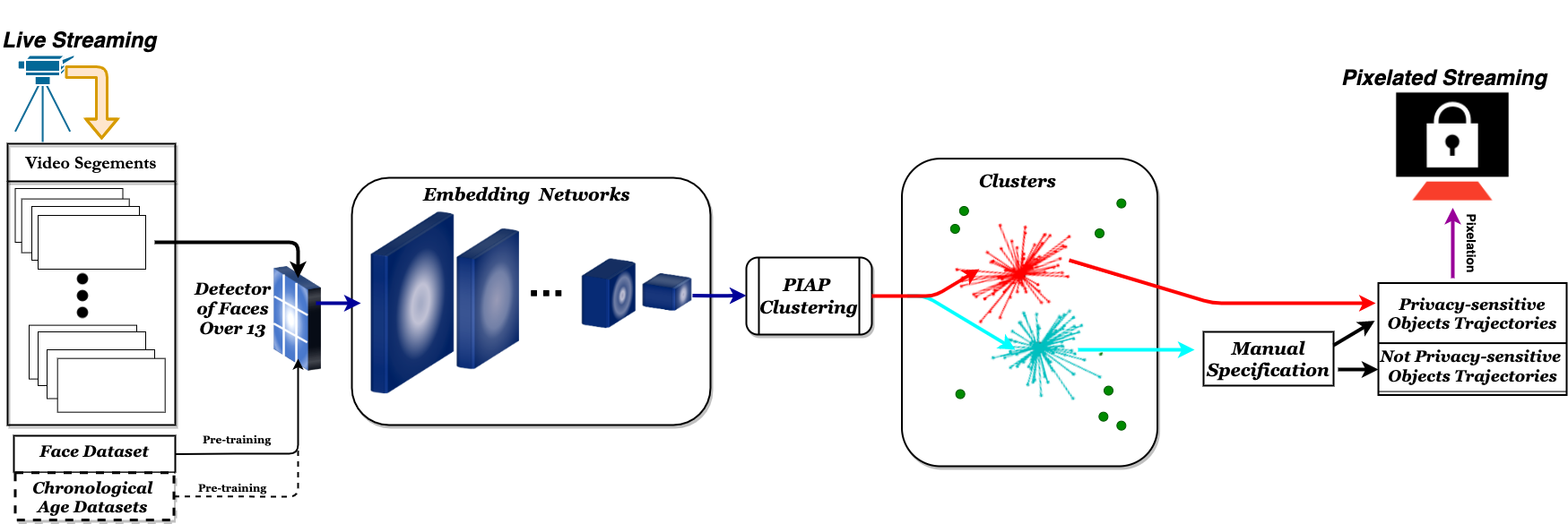}
\caption{The proposed PsOP framework.  Purple and blue lines respectively indicate the pixelation process of discriminate pixelation objects, and nonsensitive objects. Dash lines and boxes presents the extendability of PsOP through pre-training.}
%\vspace{-0.5em}
\end{figure*}

\section{Related Works}
To our best knowledge, face pixelation in live video streaming has not been well studied before. Current research of privacy-sensitive objects detection and pixelation focus on image-level~\cite{yu2016iprivacy,yu2017privacy}. \cite{yu2016iprivacy} established a multi-task learning based algorithm to extract the user-dependent privacy-sensitive objects. User blurred images are collected and fed to the multi-task CNNs for finding privacy-sensitive objects through a decision-tree shaped loss-function. Put the availability of such data in videos and accuracy aside, the relatively shallow structure along with the decision-tree shaped loss-function are unaware of the sensitiveness variations of discriminating pixelation objects in videos. The most widely applied commercial offline pixelation tools developed by YouTube Studio~\cite{youtube2020} and Microsoft Azure~\cite{Juliako2020} adopts similar detection networks in their tracking-by-detection structure. However, these works do offer us an innovative way of spotting the IPOs.
\par
As we mentioned in the introduction, current pixelation methods are incapable of functioning under poor detection. Referring to existing techniques, we firstly tried to solve the detection flaws in an end-to-end way by replacing the 2D deep CNNs with its extension 3D CNNs~\cite{ji20133d} or CNNs with attention mechanisms~\cite{li2018learning,zhu2018end}. Taking an entire video as input, 3D CNNs and the attentions consider spatial-temporal information simultaneously. Achieving great success in video segmentation~\cite{yue2015beyond,karpathy2014large}, and action recognition areas~\cite{ji20133d,tran2015learning}, they are expected to relief the detection burden. However, tested on the live streaming dataset we collected, both 3D CNNs and the attention models are quite time-costing and extremely sensitive to the video quality and length. Furthermore, without sufficient training data, as designed to extract high-level abstractions of video contexts, 3D CNNs cannot precisely handle the multi-task regression on individual frame-level. Therefore, the mosaics generated by both models always blink during live-streaming.

\par
Another proper way is to exploit multi-object tracking (MOT) algorithms that can contend with poor initiations. Although MOT are claimed to be thoroughly studied~\cite{zhang2014technology}, their tracking accuracy is still challenging in unconstrained videos. When reviewing state-of-the-art trackers, two main categories of conventional trackers can be found: offline and online trackers. The former category of trackers assumes that object detection in all frames has already been conducted, and clean tracklets are achieved by linking different detections and tracks in offline mode~\cite{berclaz2011multiple}. This property of offline trackers allows for global optimization of the path~\cite{zamir2012gmcp} but makes them inappropriate in dealing with live streaming.

\par
A typical online MOT algorithm aims at continuously locating an arbitrary target in a video with a given bounding box in the target's initial frame. State-of-the-art online MOT algorithms like KCF~\cite{henriques2014high} and ECO~\cite{danelljan2017eco} are implemented through continuous motion prediction and online learning based on the correlation filter. Varieties of tracking related works are conducted, however, they concentrate on the association of the learning-based appearance features among shots changes. Tracking problems give tacit consent to the bounding box precisions, and intrinsic detection inaccuracy is seldom mentioned or discussed. Only very few works~\cite{yu2016poi} take the detection accuracy into consideration, and they merely acknowledge the false negatives.

\iffalse

%Since the existence of discriminate pixelation objects, this cannot be solved by simply assembling a general privacy-sensitive object detector ahead of trackers. Moreover, the low recording quality and clutter backgrounds of live streaming videos jeopardize the evolutionary strategy of online-learning and aggravate the inherent drifting issue of MOT in long-term tracking. Also, conventional MOT algorithms dedicate to establish long enough trajectories during occlusion, thereby cause the over-pixelation problem in pixelation tasks. In short, even if initial information, drifting, and over-pixelation issues were fixed, without the help of PsOP in DPOs, these migrated trackers anyhow require massive human intervention to work correctly.

 %User blurred images are collected and fed to the multi-task CNNs for finding privacy-sensitive objects through a decision-tree shaped loss-function. Put the availability of such data in videos aside, the relatively shallow structure along with the decision-tree shaped loss-function are incapable of distinguishing human faces from each other or recognizing the text contents. However, these works do offer us an innovative way of spotting the IPOs.

%On video aspects, widely applied offline pixelation tools offered by YouTube are Microsoft Azure have not published their technique details. According to vast tests conducted on these tools, we firmly believe they both built on or at least profoundly effected by Multi-Object Tracking (MOT) algorithms since they behave extraordinarily close to migrated multi-object trackers.

\fi

\section{Proposed Method}
The procedure of the proposed Privacy-sensitive Objects Pixelation (PsOP) is shown in Figure 2. The process of contexts-irrelevant Indiscriminating Pixelation Objects (IPOs) are omitted since they are pixelated without generating trajectories. Live streaming videos are sliced into video segments with a fixed size for better accuracy. Then, frame-wise detections are applied through detectors pre-trained on image datasets. Detectors, along with corresponding embedding networks for every class of contexts-relevant Discriminating Pixelation Objects (DPOs) (not limited to face and text), are applied in parallel. Similarly, embeddings yielded by the same detector and embedding network are feed to the PIAP clustering. The parallel PIAP associates the same instance across frames into the same cluster, and sequential link within a cluster forms the trajectory. As not all DPOs are sensitive in a particular scene, sensitive thesaurus~\cite{bollegala2012cross} or minor manual specification are appended to filter the trajectories of DPOs further. Gaussian filters are imposed on the final trajectories for pixelation and then streaming to the audience.

\subsection{Video Segments}
The proposed PsOP leverages image-based pre-trained detection networks. Dash lines and boxes in Figure 2 shows the extendable detection of IPOs and DPOs through user-defined pre-training datasets. As false positives and negatives are common in detection, we design a buffer section at the beginning of each live streaming to promote the accuracy of trajectories without affecting the audience experience. With the PsOP's real-time efficiency, we can stack every $\mathcal{N}$ frames into a short video segment by demanding a $(2*\mathcal{N})$ frames buffering section at the very beginning of live video streaming without causing discontinuities in broadcasting.
\iffalse
\begin{proposition}
Video segments with buffering section will not cause the discontinuity in broadcasting.
\vspace{-0.3em}
\end{proposition}
\begin{proof}
The broadcasting frame per second is a constant $\{FPS\}$ during live. An arbitrary frame which is recorded at frame number $f$, will be originally broadcast to the audience at time $(\frac{f}{FPS})$.
\begin {itemize}
\item At time $\{\lceil \frac{f}{\mathcal{N}} \rceil * \frac{\mathcal{N}}{FPS}\}$ this frame is send to PsOP\\
\item PsOP takes $\{\frac{\mathcal{N}}{FPS}$\} for processing\\
\item Extra $\{\frac{f}{FPS}- \lfloor \frac{f}{\mathcal{N}} \rfloor *\frac{\mathcal{N}}{FPS}\}$ seconds needs until broadcasting
\end{itemize}
The broadcast frame number after PsOP is:\newline
\[FPS*\{\lceil \frac{f}{\mathcal{N}} \rceil * \frac{\mathcal{N}}{FPS}+\frac{\mathcal{N}}{FPS}+\frac{f}{FPS}- \lfloor \frac{f}{\mathcal{N}} \rfloor *\frac{\mathcal{N}}{FPS}\}\]
\[\Rightarrow \{f + \mathcal{N} +\lceil \frac{f}{\mathcal{N}} \rceil - \lfloor \frac{f}{\mathcal{N}} \rfloor \}    \Rightarrow f+2*\mathcal{N} \]
Therefore any frame will be broadcast with a $(2*\mathcal{N})$ time lag which can be totally covered by the buffering section.
\qedhere
\vspace{-0.5em}
\end{proof}
\fi
Video segments slice the typical hours-long live streaming into numerous segments in seconds level. Transmitted to a lag at the begging, the conflict between accuracy and efficiency is greatly alleviated. Trajectories of privacy-sensitive objects could be smoothed within and across segments. Considering the primal latency brought by data communication and video compression, such a buffering latency appended is acceptable to users.

\subsection{Indiscriminate Pixelation Objects}
Then, the Indiscriminating Pixelation Objects (IPOs) are firstly detected in video segments. We adopt the~\cite{yu2016iprivacy} as the detection network here. Training is conducted jointly on NPDI~\cite{avila2013pooling}, METU~\cite{tursun2017large}, and openALPR. The multi-task learning network is trained to handle the detection of erotic image, trademark, and plate numbers simultaneously. As in Figure 2, video segments are directly fed to the detector of indiscriminate pixelation objects colored in purple. Gaussian smooth among five consecutive frames is applied for compensating false negatives. Within a segment, the detection which has less than five bounding boxes with an overlapping ratio IOU$\leq \varepsilon$ is eliminated as false positives to avoid the blinking of mosaics. Afterward, the trajectories of indiscriminate pixelation objects are established and ready for pixelation. To avoid the domain overlapping of IPOs and DPOs, bounding boxes detected by both detectors with an IOU$> \varepsilon$ are categorized to IPOs.

\subsection{Discriminate Pixelation Objects}
To elucidate the details of proposed pixelation method for Discriminate Pixelation Objects (DPOs), faces and texts are used in the following as tangible instances. Following the embedding then clustering pixelation process for DPOs, in this paper, we use MTCNN~\cite{zhang2016joint}, and CosFace~\cite{wang2018cosface} to process every frame in turn. The application of MTCNN and CosFace considers the convenience for latter cosine similarity in PIAP. The detection and embedding networks could be substituted by other state-of-the-art methods. MTCNN accepts arbitrary size inputs and detects face larger than $12*12$ pixel. CosFace generates $512$ dimension feature vector for each detected face. Faces are aligned to the frontal pose through the affine transformation before embedding. Similarly for texts, Textboxes~\cite{liao2017textboxes} and FastText~\cite{joulin2016fasttext} are used in texts detection and embedding. TextBoxes receives input images containing texts with an arbitrary scale. A bit close to MTCNN, frames are rescaled to $300*300$ for efficiency, and Non-Maximum Suppression (NMS) are also leveraged to boost the accuracy. FastText is a C-BOW like fast text classification network, and the result of the second last layer before softmax regression is extracted as embedding. The embedding dimension of FastText is defined to $128$. The clustering for faces and texts is conducted in parallel.
%Extracted texts are further filtered through sensitive thesaurus based on~\cite{bollegala2012cross}. Similar trajectory smoothing strategy as IPOs is applied for the texts attributed to sensitive objects.

\par
The proposed PIAP is then activated to connect the same faces or the same piece of text across frames according to face or text vectors. Do notice that the efficacy of PIAP remains the same in establishing trajectories for other objects of this type. DBSCAN~\cite{ester1996density} and Affinity Propagation (AP)~\cite{frey2007clustering} are the candidates under unpredictable cluster numbers. As noisy detection results are common in videos and the data-size for clustering is also unbalanced, the density-based DBSCAN is excluded. To revise the false detection, variation of embedded vectors, and boost the clustering speed, we employ the position information and incremental clustering in the proposed PIAP. Sequential links within each cluster form the trajectory of each face and text, and likewise, Gaussian smooth is applied within a segment before pixelation.
\subsubsection{Affinity Propagation (AP)}
Similar to traditional clustering algorithms, the first step of classic AP is the measurement of the distance between data nodes, denoted as the similarities. Following the common notation in AP, $i$ and $k$ ($i,k\in R^D$, $D=512$ for face and $D=128$ for text) are two of the data nodes. $S$ is the similarity matrix stores the similarities between every two nodes. $S(i,k)$ is the element on row $i$, column $k$ of $S$ and $S(i,k)$ denotes the similarity between data node ${i}$ and ${k}$, thereby indicating how well the data node $k$ is suited to be the exemplar for data point $i$. Similar notations are also used in below.
\par
The core of AP is that a series of responsibilities $R(i,k)$ and availabilities $A(i,k)$ messages are passed among all data nodes to reach the consensus on exemplars' selection. $R$ and $A$ are the responsibility and availability matrix. $i$ passes $R(i,k)$ to its potential exemplar $k$ indicating the current willingness of $i$ choosing $k$ as its exemplar considering all the other potential exemplars. Correspondingly, $k$ responds $A(i,k)$ to $i$ update the current willingness of $k$ accepting $i$ as its member considering all the other potential members. The sum of $R(i,k)$ and $A(i,k)$ can directly give the fitness for choosing $k$ as the exemplar of $i$. The consensus of the sum-product message passing process ($R(i,k)$ and $A(i,k)$ remain the same after iterations) stands for the final agreement of all nodes in the selection of exemplars and the association of clusters is reached. Apart from the ill-defined cluster number, AP is not sensitive to the initialization settings; selects real data nodes as exemplars; allows asymmetric matrix as input; is more accurate when measured in the sum of square errors. Therefore, considering the subspace distribution (least square regression),  AP is effective in generating robust and accurate clustering results for high dimension data like face vectors.

\par
The governing equations for message passing in AP are:

\begin{equation}
R(i,k) \leftarrow S(i,k) - \max_{{k'}, s.t.{k'} \neq k}\{A(i,k')+S(i,{k'})\}
\end{equation}
\begin{equation}
A(i,k) \leftarrow \min\bm{\{} {0, R(k,k)+ \sum_{{i'},{i'}\notin {\{i,k\}}} \max \{0, R({i'},k)\} \bm{\}}}
\end{equation}
Equation (3) is used to fill in the elements on the diagonal of the availability matrix:
\begin{equation}
A(k,k) \leftarrow \sum_{i', s.t. i' \neq k}\max\{0, R(i',k) \}
\end{equation}
Update responsibilities and availabilities according to (1), (2) and (3) till convergence, then the criterion matrix $C$ which holds the exemplars is the sum of the $A$ and $R$ at each location.
\begin{equation}
C(i,k) \leftarrow R(i,k)+A(i,k)
\end{equation}
The highest value of each row of the criterion matrix is designated as the exemplar.

%\section{Proposed Method}
\subsubsection{Positioned Incremental Affinity Propagation (PIAP)}
Clustering builds the connection of the same face across frames since its result could be facilely corrected with some intervention. With the results of face detection and recognition, the proposed PIAP process in a segment-wise way to group the faces within and across segments simultaneously. Apparently, the longer segments offer more information but also incur more noise and reduce efficiency. In this paper, the proposed PsOP cuts every 150-frames into a segment and leverages the whole existing context to reach accurate and fast face pixelation. Define a data stream ${Z=\{Z_1,Z_2,...,Z_n\}}$ is sequentially collected face/text feature vectors of the current video segment. i.e. $Z$ is an $512*n$ matrix with $512$ dimension feature vector.
\par
For the normalization in the PIAP, cosine similarity is brought for measurements. As an instance of DPOs (like a person's face) can only appear at one position in a single frame, instances that belong to the same frame are set to the minimum similarity value $-1$. Let $j$ stands for the other instances that belong to the same frame as $i$. The similarity matrix can be generated as:
\begin{equation}
   S(i,k)=
    \left\{
             \begin{array}{lr}
             \frac{i \cdot k}{\Vert i \Vert \Vert k \Vert} -1,  \quad\mbox{ if }\ k \notin{j}\\

          \\
             -1, \quad \mbox{ if } \ k \in j
             \end{array}
    \right.
\end{equation}
\par
Moreover, in (2), the left side of the equation is the accumulated evidence of how much $k$ can stand for itself ($R(k,k)$), and how many others $k$ is also responsible to stands for ($\sum_{{i'},{i'}\notin {\{i,k\}}} \max\{0, R({i'},k)\}$). Every positive responsibilities of $k$ contributes to $A(i,k)$. Therefore, according to (5), $A(i,k)$ should not accumulate $j$'s choice as the evidence for representing $i$. One step further, the choice of $j$ will actually repel $i$ in availability, and the message passing of $A(i,k)$ shall be rewritten as:
\begin{equation}
    \begin{aligned}
      &A(i,k) \leftarrow  \min\bm{\{}0, R(k,k)+ \\
             &\sum_{{i'},{i'}\notin {\{i,j,k\}}} \max\{0, R({i'},k)\}   -\sum_{{i'},{i'}\in {\{j\}}} \max\{0, R({i'},k)\}\bm{\}}
           \end{aligned}
\end{equation}
(5), (6) ensures $i$ and $j$ are mutual exclusive in the clustering process.
\par
After the revision of the position information, the challenge for extending AP into an incremental way is that the data nodes received at different time\-stamps stay at varying statuses with disproportionate responsibilities and availabilities value.
\par
However, video frames are strongly self-correlated data. We could assign a proper value to newly-arrived vectors according to the ones in the previous segment without affecting the clustering purity. The embedded vectors of a particular person or text shall stay close with each other in the feature space across different frames. Thus, our incremental AP algorithm is proposed based on the fact that if two detected faces/texts are in adjacent segments and refer to the same person/the same piece of texts, they should not only be clustered into the same group but also have the same responsibilities and availabilities. Such fact is not well considered in past studies of incremental affinity propagation~\cite{sun2014incremental,wang2013multi}.
\par
Following the common notations in AP, the similarity matrix is denoted as $S_{t-t'}$ at time $t-t'$ with ($M_{t-t'}*M_{t-t'}$) dimension where $(t-t')= \frac{\mathcal{N}}{FPS}$. And the responsibility matrix and availability matrix at time $t-t'$ are $R_{t-t'}$ and $A_{t-t'}$ with a same dimension as $S_{t-t'}$.
Then, the update rule of $R_{t}$ and $A_{t}$ respect to $R_{t-t'}$ and $A_{t-t'}$ can be written as:
\begin{equation}
  %\vspace{-1em}
    R_{t}(i,k)=
    \left\{
             \begin{array}{lr}
             R_{t-t'}(i,k), \quad i\leq M_{t-t'}, k\leq M_{t-t'}\\
             \\
             R_{t-t'}(i',k), \quad i > M_{t-t'}, k\leq M_{t-t'}\\
             \\
             R_{t-t'}(i,k'), \quad i\leq M_{t-t'}, k > M_{t-t'}\\
             \\
             0, \quad i > M_{t-t'}, k > M_{t-t'}
             \end{array}
\right.
%\vspace{-1em}
\end{equation}
Note that the dimension of three matrices is increasing with time. $R_{t}(i,k)$ is the newly arrived face vectors of a segment at time $t$. $M_{t-t'}$ stands for the amount of faces at time $t-t'$.
\begin{equation}
    i'=\arg \max_{i', i'\leq M_{t-t'}} \left \{S(i,i')\right \}
\end{equation}
Easily, the $A_{t}$ could be updated through
\begin{equation}
  %\vspace{-1em}
    A_{t}(i,k)=
    \left\{
             \begin{array}{lr}
             A_{t-t'}(i,k), \quad i\leq M_{t-t'}, k\leq M_{t-t'}  \\
             \\
             A_{t-t'}(i',k), \quad i > M_{t-t'}, k\leq M_{t-t'}\\
             \\
             A_{t-t'}(i,k'), \quad i\leq M_{t-t'}, k > M_{t-t'}\\
             \\
             0, \quad i > M_{t-t'}, k > M_{t-t'}
             \end{array}
    \right.
    %\vspace{-1em}
\end{equation}

\par
Denote $z_{p}^{q}=\{z_{1}^{q},z_{2}^{q},z_{3}^{q}...z_{p}^{q}\}$ as the set of all $p$ vectors extracted in segment $q$. The full process of PIAP algorithm can be summarized as Algorithm 1.

\begin{algorithm}[htbp]
\begin{flushleft}
\caption{Positioned Incremental Affinity Propagation}
\label{alg:algorithm}
\textbf{Input}: $R_{t-t'}$,$A_{t-t'}$,$C_{t-t'}, z_{p}^{q}$\\
\textbf{Output}: $R_{t},A_{t},C_{t}$
\end{flushleft}
\begin{algorithmic}[1] %[1] enables line numbers
\WHILE {$q$ is not end of a live-streaming}
\STATE Compute similarity matrix according to (5).
\IF {$q$ is the first video segment of a live-stream}
\STATE Assign zeros to all responsibilities and availabilities.
\ELSE
\STATE Compute responsibilities and availabilities for $z_{p}^{q}$ according to equation (7), (8) and (9).
\STATE Extend responsibilities matrix $R_{t-t'}$ to $R_{t}$, and availabilities $A_{t-t'}$ to $A_{t}$.
\ENDIF
\STATE Message-passing according to equation (1), (6) and (3) until convergence.
\STATE Compute exemplars and clustering results $C_t$ as equation (4).
\ENDWHILE
\end{algorithmic}
\end{algorithm}

\par
Handled by PIAP, faces, texts, and alike DPOs could be intra-class distinguished. Cluster results are smoothed similar as the IPOs for the exclusion of false negatives and positives. Faces and texts within a cluster are then linked sequentially to form a trajectory.
\subsection{Pixelation on Trajectories of Privacy-sensitive Objects}
In live video streaming, the appeared non-streamers are much more than streamers but own a far smaller sample-size for each non-streamer. Thus, as in the rightmost part of Figure 1, with trajectories built by PIAP, manual specification (like screen touching) is involved solely on the face of a newly appeared streamer. Also, texts are further filtered through checking the sensitive thesaurus based on~\cite{bollegala2012cross}.  With the rolling of PIAP, the previously specified cluster is capable of guiding the future aggregation of the same faces or texts.
\par
The trajectories of the faces of the non-streamers, the sensitive texts along with IPOs are gathered as the trajectories of privacy-sensitive objects. Gaussian filters are applied on these trajectories for pixelation before broadcasting to the audience.

\section{Experiments and Discussion}

\begin{table*}[h]
\caption{Details of the live video streaming dataset}
\centering
\begin{tabular}{|c||cccc||cccc|}
\hline
{Dataset} &\tabincell{c}{Quantity\\ of videos} & Category  & Resolution & \tabincell{c}{People occurred}  &Frames & \tabincell{c}{Privacy-sensitive \\objects Labels*} &{\tabincell{c}{IPOs\\Labels*}}   &{\tabincell{c}{DPOs\\Lables*}}  \\
\hline\hline
$HS$    &4 &  a,b,c,d       & 720p/1080p   &$\gg$ 2     &4133  &27337 &6255 &21082 \\

$LS$    &8 &  a,b,c,d      & 360p/480p   & $\gg$ 2    &4680  &23869  &3975 &19894  \\

$LN$     &4 &  a,b,c,d       &360p/480p   & $\le$2    &5692 &16994  &7946 &9048\\

$HN$    &4 &  a,b,c,d      &720p       & $\le$2    &4867  &16061  &10805 &5256\\
\hline
\end{tabular}
\begin{tablenotes}
      \small
      \item $\quad\quad\quad$*: the value amounts the occurrences of the relative objects. Meaning the same object is repeatedly counted in different frames.
    \end{tablenotes}
%\label{tab:plain}
%\vspace{-1em}
\end{table*}

\begin{table*}[h]
\centering
\caption{Pixelation results on collected live video streaming dataset}
\begin{tabular}{|c||cccc||cccc||c|c|c|}
\hline
\multicolumn{1}{|c||}{\multirow{2}{*}{Method}} &{SOPA$\uparrow$}  &{SOPA$\uparrow$}  &{SOPA$\uparrow$}  &{SOPA$\uparrow$} &{SOPP$\uparrow$}  &{SOPP$\uparrow$} &{SOPP$\uparrow$}
	\multirow{2}{*}&{SOPP$\uparrow$}
    &\multicolumn{3}{c|}{Entire Dataset}\\
    \cline{10-12}
    &($HS$)&($LS$)&($HN$)&($LN$)&($HS$)&($LS$)&($HN$)&($LN$)&
    \tabincell{c}{MP$\uparrow$\\(frames)} &{OPR$\downarrow$} &{FPS$\uparrow$} \\
\hline \hline
YouTube~\cite{youtube2020}     &0.61  &0.57 &0.68 & 0.57 &0.70 &0.64 &0.78 &0.63 &238   & 0.44   & $N/A$\\
Azure~\cite{Juliako2020}      &0.63  &0.57 &0.66  &0.56 &0.71 &0.66 &0.77 &0.66   &203  & 0.43    & $N/A$  \\
KCF~\cite{henriques2014high}     &0.59 &0.52 &0.68 & 0.61  &0.69 &0.63 &0.79 &0.70 &166  &0.47  &\textbf{$>$100} \\
ECO~\cite{danelljan2017eco}        &0.57 &0.61 &0.69 & 0.66  &0.66 &0.75 &0.79 &0.74    &178  &0.41 &{30\textasciitilde 50} \\
\hline\hline
%\tabincell{c}{PsOP\\(Pyrimid+ArcFace)} & \textbf{0.73}  &0.67  &\textbf{0.75} &\textbf{0.71} &\textbf{0.84} &0.77 &0.88 &\textbf{0.88} &362  &\textbf{0.21}  &15\textasciitilde30\\
\tabincell{c}{PsOP} & \textbf{0.73}  &\textbf{0.69}  &\textbf{0.75} &\textbf{0.71}  &\textbf{0.80} &\textbf{0.79} &\textbf{0.84} &\textbf{0.83} &\textbf{387} &\textbf{0.21} &{18 \textasciitilde 30} \\
\hline
\end{tabular}
\begin{tablenotes}
      \small
      \item $\quad\quad$"$\uparrow$"$\&$"$\downarrow$"stand for the higher the better and the lower the better respectively.
    \end{tablenotes}
%\label{tab:plain}
%\vspace{-1em}
\end{table*}

%%%%%%%%%%%%%%%%%%%%%%%%%%%%%%%%%%%%%%%%%%%%%
%%%%%%%%%%%%%%%%%%%%%%%%%%%%%%%%%%%%%%%%%%%%%

\begin{table*}[h]
\centering
\caption{Pixelation results of DPOs on live video streaming dataset}
\begin{tabular}{|c||cccc||cccc||c|c|}
\hline
\multicolumn{1}{|c||}{\multirow{2}{*}{{Method}}} &{SOPA$\uparrow$}  &{SOPA$\uparrow$}  &{SOPA$\uparrow$}  &{SOPA$\uparrow$} &{SOPP$\uparrow$}  &{SOPP$\uparrow$} &{SOPP$\uparrow$}
	\multirow{2}{*}&{SOPP$\uparrow$}
    &\multicolumn{2}{c|}{Entire Dataset}\\
    \cline{10-11}
    &($HS$)&($LS$)&($HN$)&($LN$)&($HS$)&($LS$)&($HN$)&($LN$)&
    \tabincell{c}{MP$\uparrow$\\(frames)} &{OPR$\downarrow$}\\
\hline \hline
YouTube~\cite{youtube2020}     &0.45  &0.42 &0.56 & 0.49 &0.53 &0.47 &0.77 &0.63 &238 &0.56               \\
Azure~\cite{Juliako2020}      &0.43  &0.47 &0.54  &0.53 &0.50 &0.53 &0.70 &0.68   &203                 &0.54\\
KCF~\cite{henriques2014high}       &0.35 &0.32 &0.38 & 0.31  &0.41 &0.40 &0.44 &0.40    &113               &0.64\\
ECO~\cite{danelljan2017eco}         &0.27 &0.28 &0.34 & 0.31  &0.37 &0.39 &0.41 &0.40    &148
     &0.59\\
\hline\hline
\tabincell{c}{PsOP\\(MTCNN+CosFace)} & \textbf{0.63}  &0.60  &\textbf{0.71} &\textbf{0.66}  &\textbf{0.80} &0.77 &\textbf{0.86} &\textbf{0.85} &362 &\textbf{0.34} \\
\hline
\tabincell{c}{PsOP\\{(Pyramid+ArcFace)}}  &\textbf{0.63}  &\textbf{0.62}  &\textbf{0.71} &0.65  &\textbf{0.80} &\textbf{0.79} &\textbf{0.86} &\textbf{0.85}&387 &\textbf{0.34} \\
\hline
\end{tabular}
\begin{tablenotes}
      \small
      \item $\quad\quad$"$\uparrow$"$\&$"$\downarrow$"s.tand for the higher the better and the lower the better respectively.
    \end{tablenotes}
%\label{tab:plain}
%\vspace{-1em}
\end{table*}
%\begin{tabular}{|c||cccc||c|}
%\toprule
%Method  &\tabincell{c}{MFPA\\on $HS$}  &\tabincell{c}{MFPA\\on $LS$}  %&\tabincell{c}{MFPA\\on $HN$}  &\tabincell{c}{MFPA\\on $LN$} %&OPR\\
%\hline

%YouTube     &0.45  &0.42 &0.56 & 0.49  &0.56   \\
%Azure      &0.43  &0.47 &0.54  &0.53   &0.54 \\
%KCF        &0.35 &0.32 &0.38 & 0.31      &0.64\\
%ECO        &0.27 &0.28 &0.34 & 0.31     &0.59 \\

\subsection{Dataset}
As the privacy-sensitive objects pixelation problem in live video streaming has not been studied before, there is no available dataset or benchmark tests for reference. We collected and built a video live streaming video dataset from YouTube and Facebook platforms and manually labeled the sensitive objects for the experiments. In total, 20 streaming videos come from 4 different streaming categories (a:street interview; b:street streaming; c:flash activities; d:dancing) of 19372 frames are manually labeled and will be available to public. As demonstrated in Table 1, these videos are further divided into four groups according to the broadcasting resolution, and the number of people showed up in streaming. Live streaming videos with at least 720p resolution are marked as high-resolution $H$, and the rest are low-resolution $L$. Similarly, the videos contain more than two people are sophisticated scenes $S$, and the rest are naive ones $N$.The number of labeled bounding boxes regarding privacy-sensitive objects, IPOs, and DPOs are also listed to exhibit the dense annotation.

\subsection{Parameters}
All the parameters remain unchanged for entire live-streaming video tests. $\mathcal{N}=150$ for every segment contains 150 frames. 10 seconds for buffering as $2*\mathcal{N}/30$. CosFace resizes the cropped face to $112*96$; weight decay is 0.0005. The learning rate is 0.005 for initialization and drops at every 60K iterations. The damping factor for PIAP is default 0.5 as it in AP. Detection threshold for MTCNN is [0.7,0.8,0.9]. The IOU rate $\varepsilon$ is [0.7].

\subsection{Evaluation Metrics}
Setting our manual pixelation annotations in the collected video live streaming dataset as the baseline, we migrated and adopt the algorithms including ECO and KCF tracker, currently applied commercial tools like YouTube Studio and Microsoft Azure, and the proposed PsOP for comparison. Sensitive Objects Pixelation Accuracy (SOPA), Sensitive Objects Pixelation Precision (SOPP), and Most Pixelated (MP) frames are the most concerned metrics. Specifically:
\begin{equation}{\nonumber}
SOPA=1- {\frac{\sum_{t}(m_t+fp_t+mm_t)}{\sum_{t} {g_t}}}
\end{equation}
where $m_t$, $fp_t$, $mm_t$, $g_t$ correspond to the missed pixelation, false positives in pixelation, missmatched pixelation, total pixelation in frame $t$. SOPA reflects the overall performance of the PsOP.
\begin{equation}
\nonumber
SOPP={\frac{\sum_{i,t}d_{i,t} }{\sum_{t} {c_t}}}
\end{equation}
where $d_{i,t}$ is the bounding box overlapping ratio of pixelated sensitive object $i$ with its labelled ground truth. $c_t$ is the total number of matched pixelation in frame $t$. Therefore, SOPP states the degree of drifting and pixelation precision. Over-Pixelated Ratio (OPR) is another important metric indicating the degree of over-pixelation problem.
\begin{equation}\nonumber
OPR=\sum_{t}( fp_t )/ {\sum_{t} {c_t}}
\end{equation}

\subsection{Ablation Study}
In Table 2, all the algorithms get their best performance on the High-resolution Naive-scene ($HS$) sub-dataset.KCF and ECO here are optimized for privacy object tracking tasks with detection based calibration. To conduct the comparisons, we activated manual specification on occurrences of sensitive objects for the first four algorithms. However, affected by the noisy detection results, KCF styled online learning algorithms are maladjusted in video live streaming. Although ECO applies higher dimension features than KCF, the SOPA of ECO is lower than KCF. Overall, our PsOP achieves better performance in terms of SOPA, SOPP, MP, and OPR. Particularly, PsOP noticeably increases most pixelated frames and decrease the over-pixelation ratio on the entire dataset, indicating that PsOP can construct more robust trajectories and avoid pixelation if the object is temporally not identifiable. The high value in SOPP suggests the trajectory drifting in PsOP is relatively rare or happens at a later time.

\par
Table 3 shows the pixelation results solely on DPOs. Also, manual specification is activated for the first four methods. PsOP gains a surpassing pixelation accuracy and proves the proposed embedding and PIAP clustering framework is sharp in dealing with DPOs. Listed at the bottom line of Table 3, the boxed PsOP substitutes the detection and embedding network with more advanced PrimidBox~\cite{tang2018pyramidbox}, ArcFace~\cite{deng2019arcface}. The last section shows the network selection is actually not that important for PIAP, as long as they are deep architectured, well-tunned networks.
\par

\begin{table}[h]
\centering
\caption{Clustering purity and efficiency for faces}
\begin{tabular}{|c|c|c|c|c|}
\hline
{Dataset} &{Method} &{Purity$\uparrow$} &\tabincell{c}{Cluster nubmer\\ (Clustered/Truth)}   &{Time(s)$\downarrow$}        \\
\hline\hline
\multirow{3}{*}{$HS$}
& {AP}  & {0.81} & {19/17}  & {1.82}   \\
\cline{2-5}
& {PAP}  & {0.89}  &{17/17}& {1.82}   \\
\cline{2-5}
& {PIAP}  & {0.89} & {17/17} & {0.07}   \\
\hline\hline
\multirow{3}{*}{$LS$}
& {AP}  & {0.72}  & {37/43}& {3.05}   \\
\cline{2-5}
& {PAP}  & {0.84} & {41/43}& {3.05}   \\
\cline{2-5}
& {PIAP}  & {0.83} & {41/43}& {0.13}   \\
\hline\hline
\multirow{3}{*}{$HN$}
& {AP}  & {0.86} & {6/6} & {0.75}   \\
\cline{2-5}
& {PAP}  & {0.93} & {6/6}& {0.75}   \\
\cline{2-5}
& {PIAP}  & {0.93}& {6/6} & {0.04}   \\
\hline\hline
\multirow{3}{*}{$LN$}
& {AP}  & {0.89} & {7/7}& {0.80}   \\
\cline{2-5}
& {PAP}  & {0.92}& {7/7} & {0.80}   \\
\cline{2-5}
& {PIAP}  & {0.91}& {7/7} & {0.03}   \\
\hline
\end{tabular}
%\label{tab:plain}
%\vspace{-1em}
\end{table}

\begin{figure}[h]
\begin{center}$
\begin{array}{ccc}
\vspace{-0.2em}
\includegraphics[width=28.4mm]{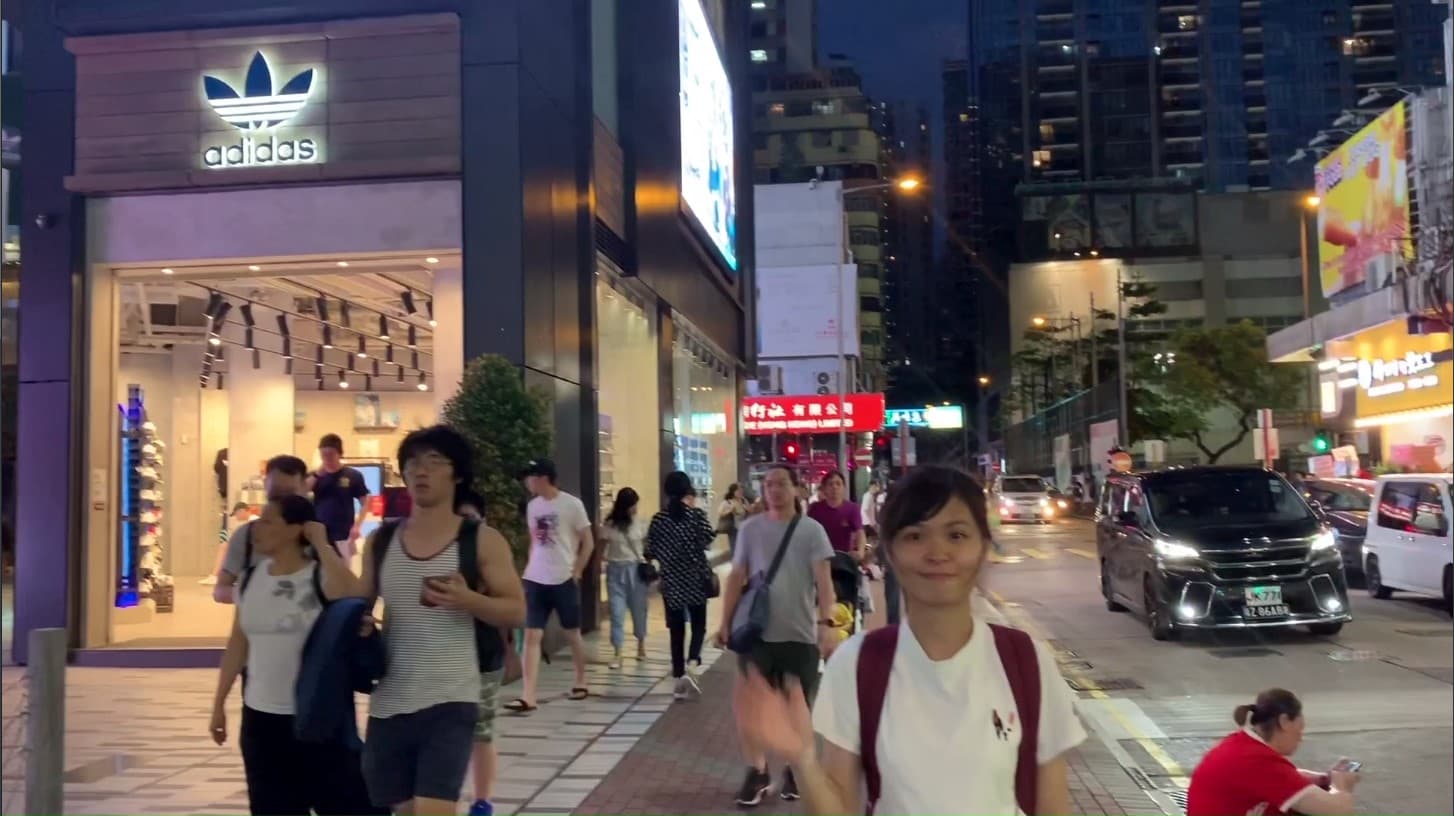}& \hspace{-0.8em}
\includegraphics[width=28.4mm]{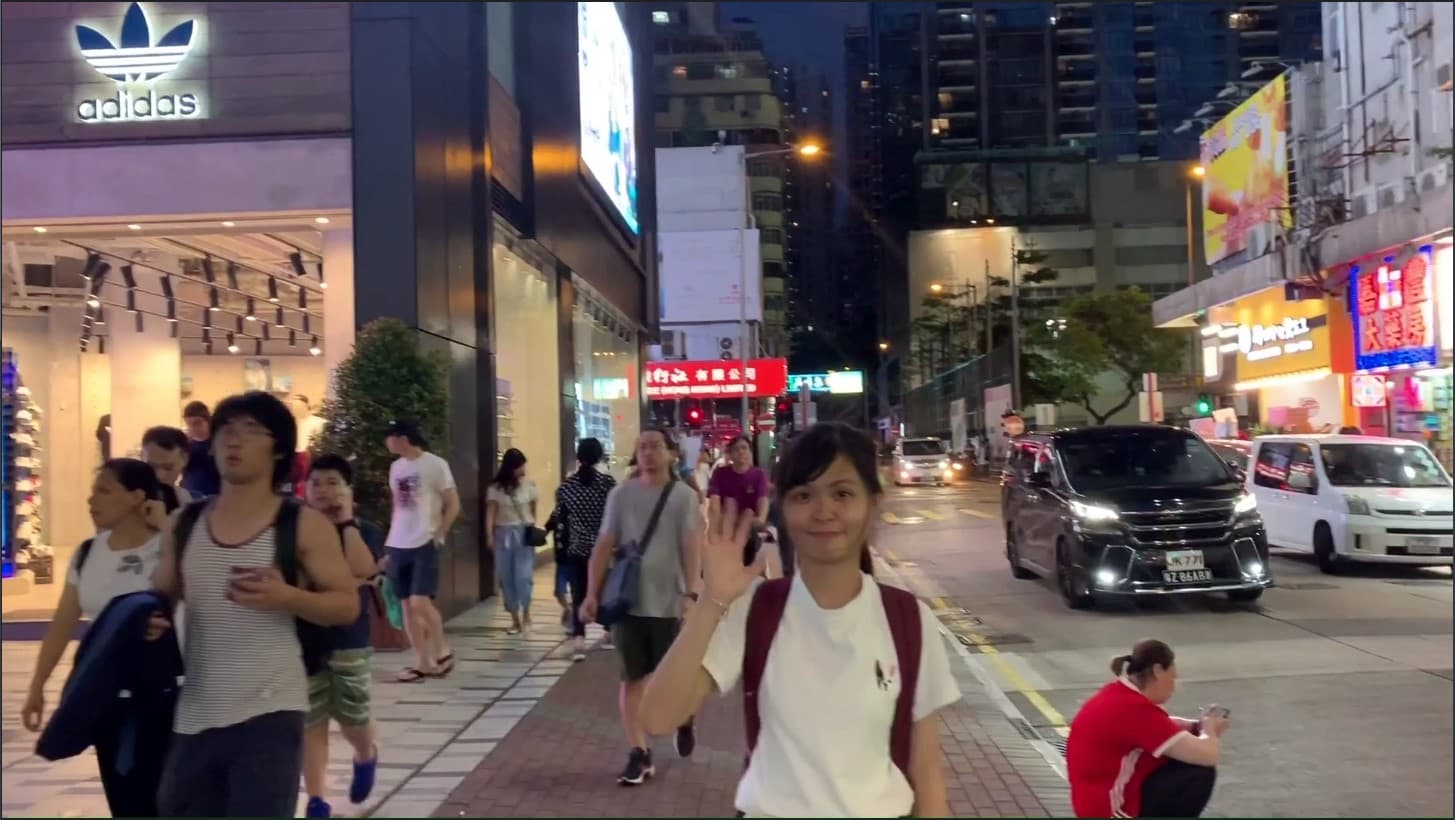}& \hspace{-0.8em}
\includegraphics[width=28.4mm]{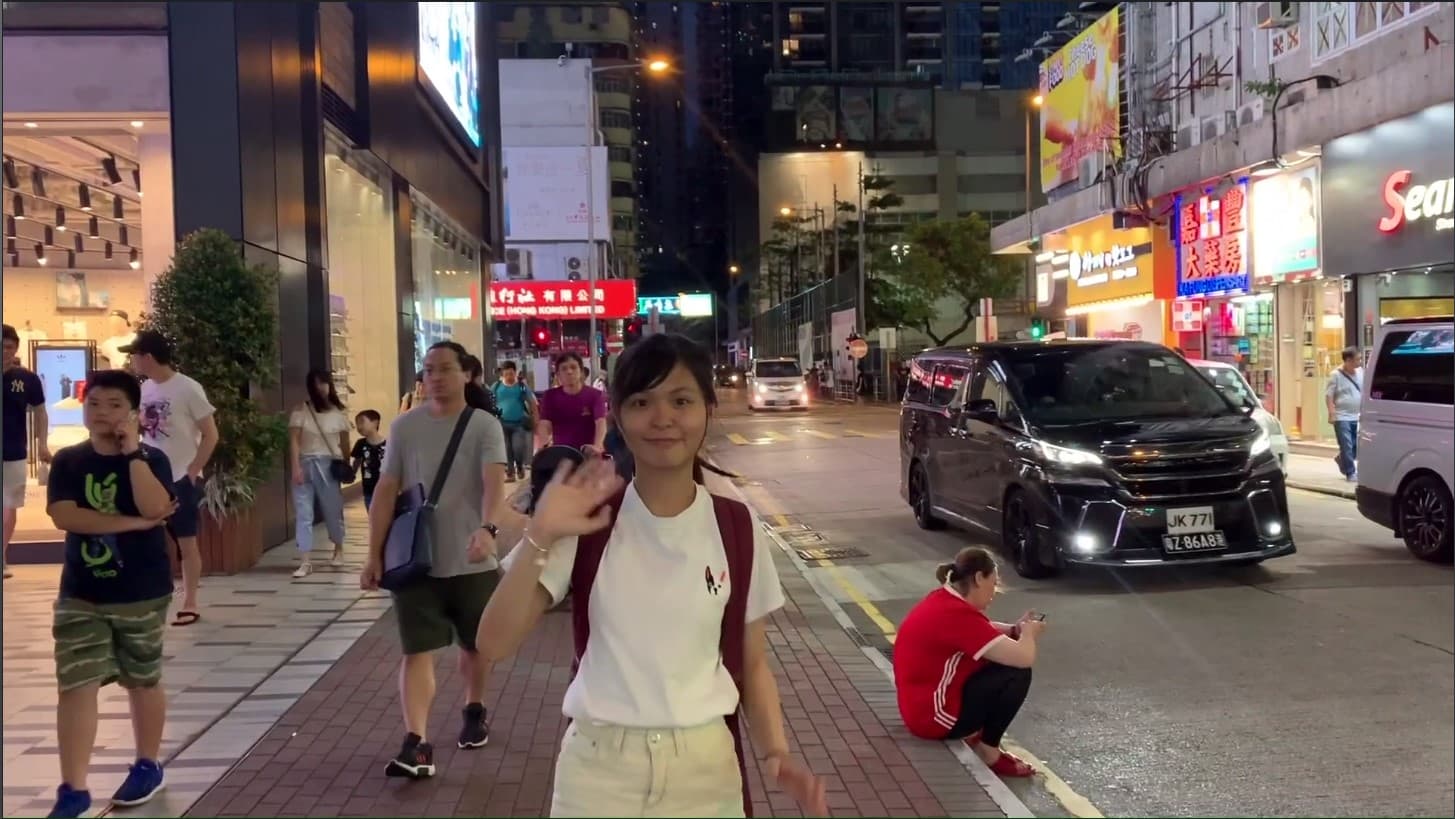} \\
\vspace{-0.2em}
\includegraphics[width=28.4mm]{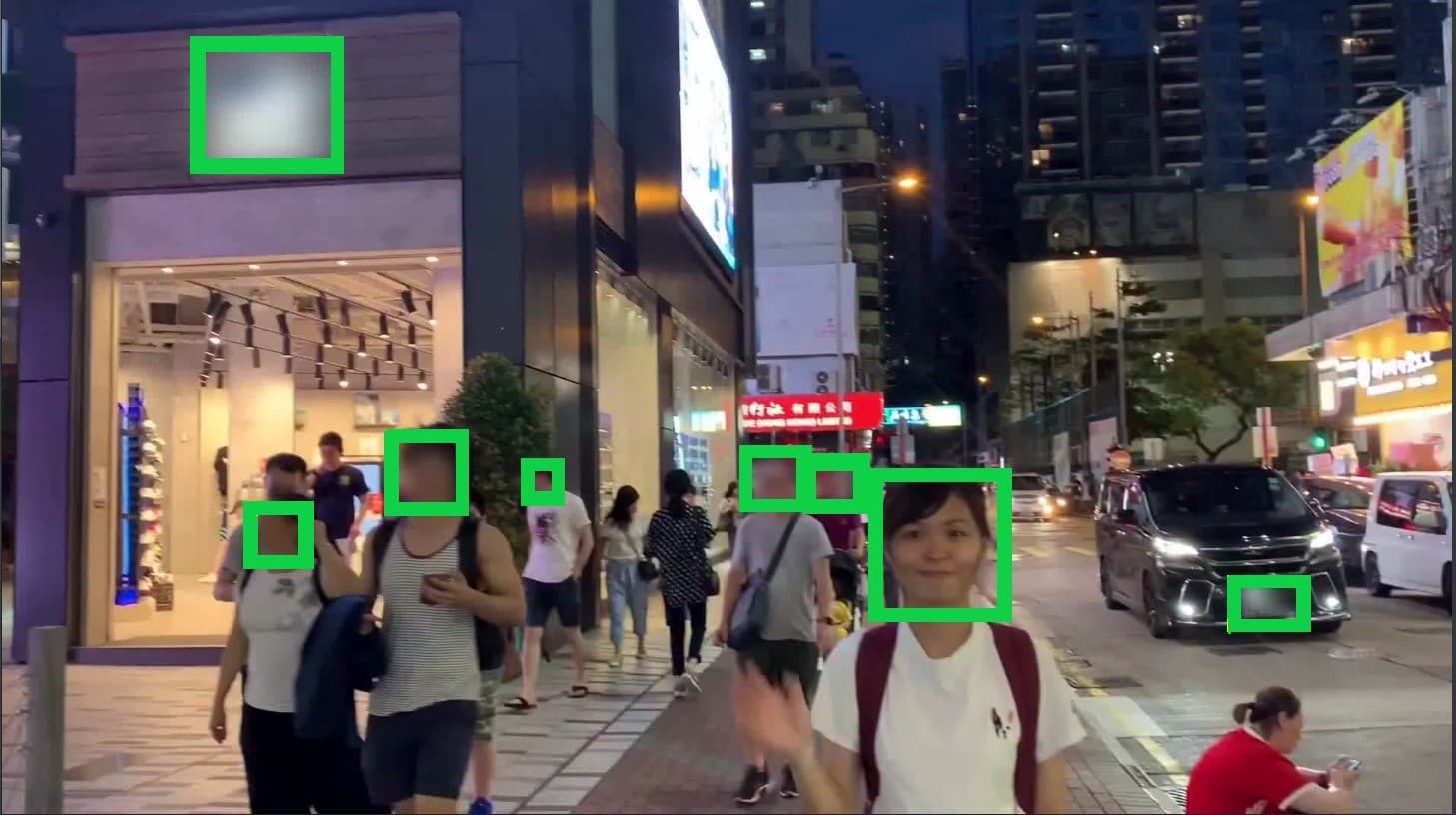}& \hspace{-0.8em}
\includegraphics[width=28.4mm]{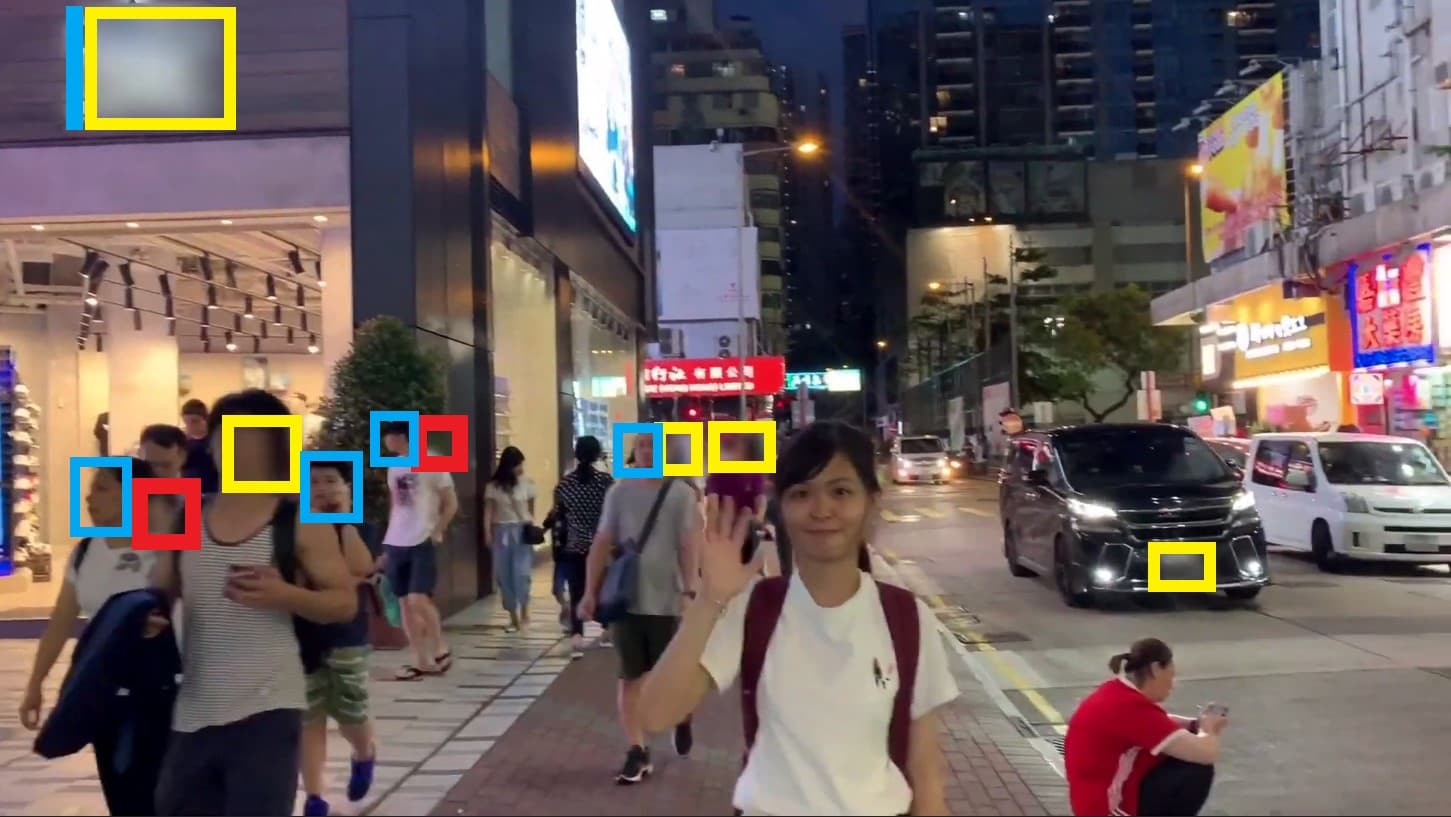}& \hspace{-0.8em}
\includegraphics[width=28.4mm]{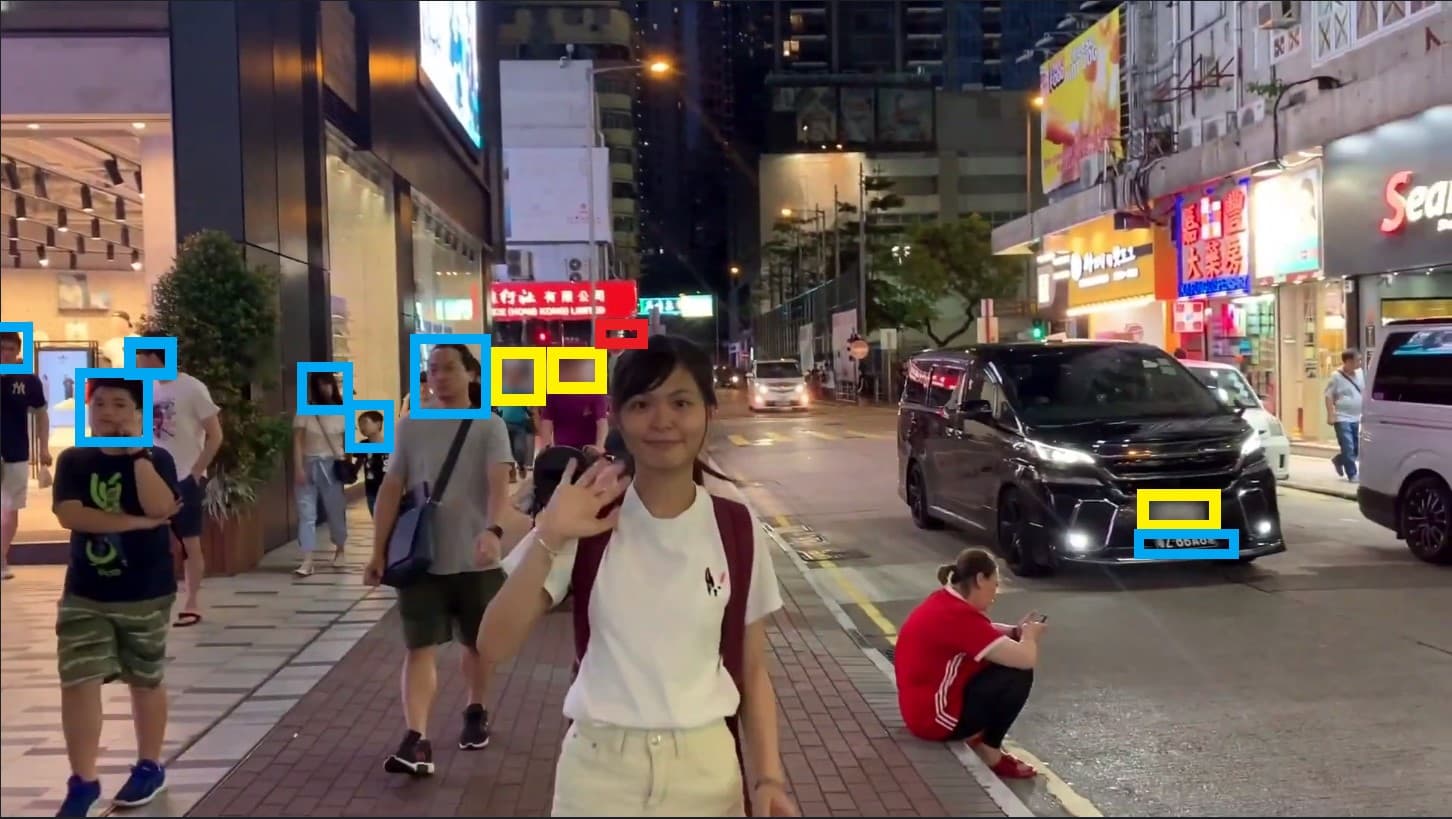}\\
\vspace{-0.2em}
\includegraphics[width=28.4mm]{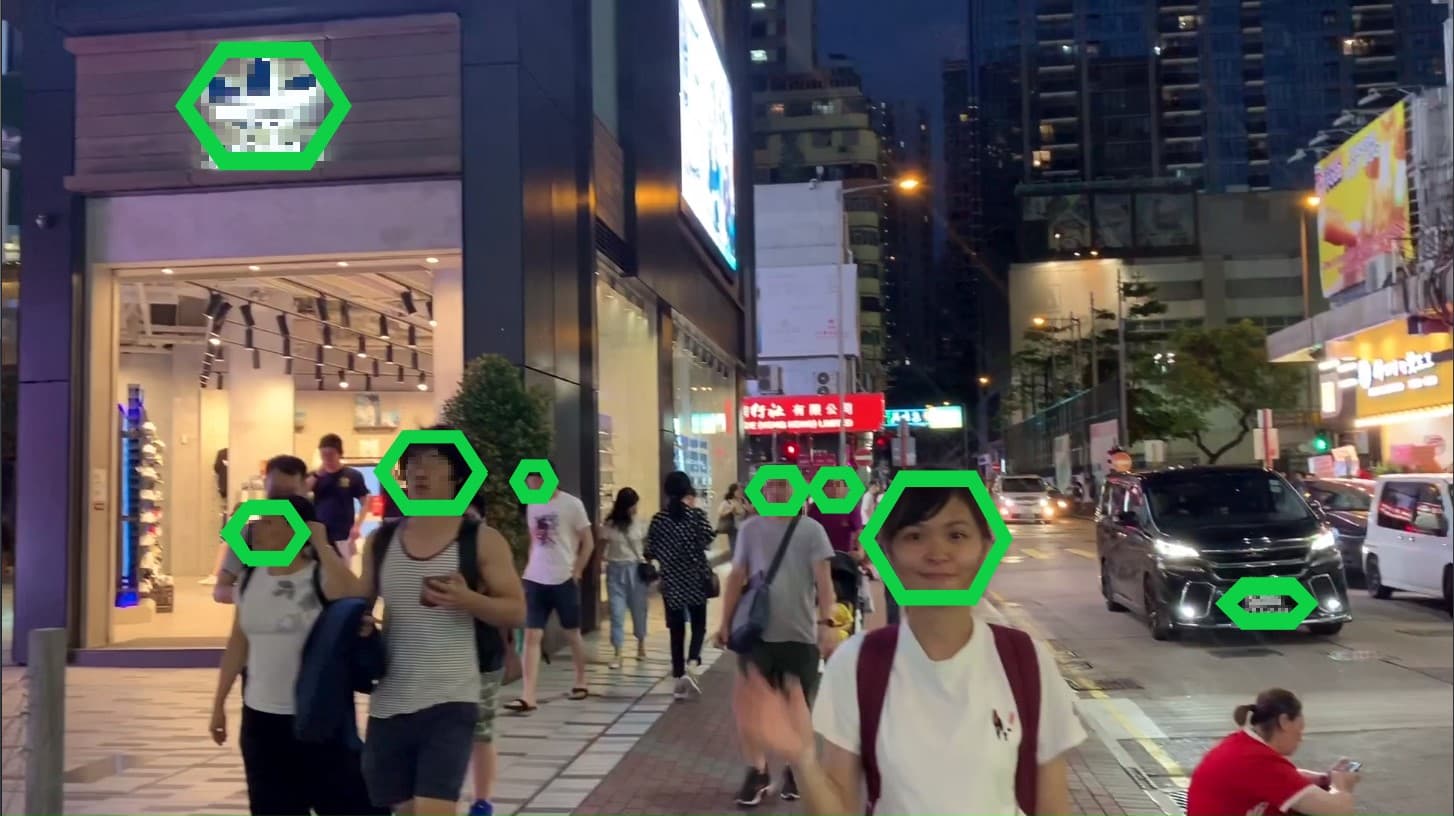}& \hspace{-0.8em}
\includegraphics[width=28.4mm]{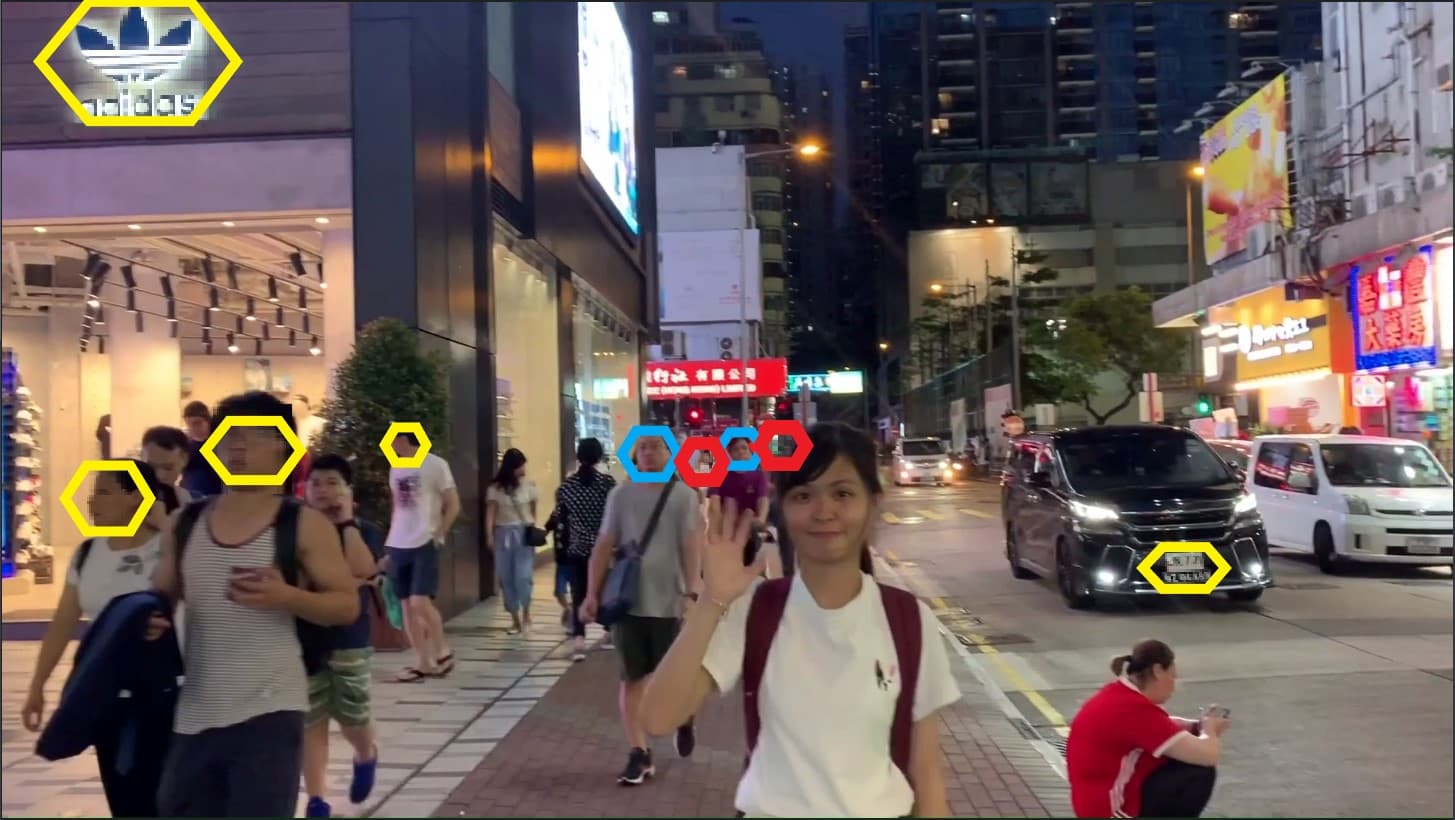}& \hspace{-0.8em}
\includegraphics[width=28.4mm]{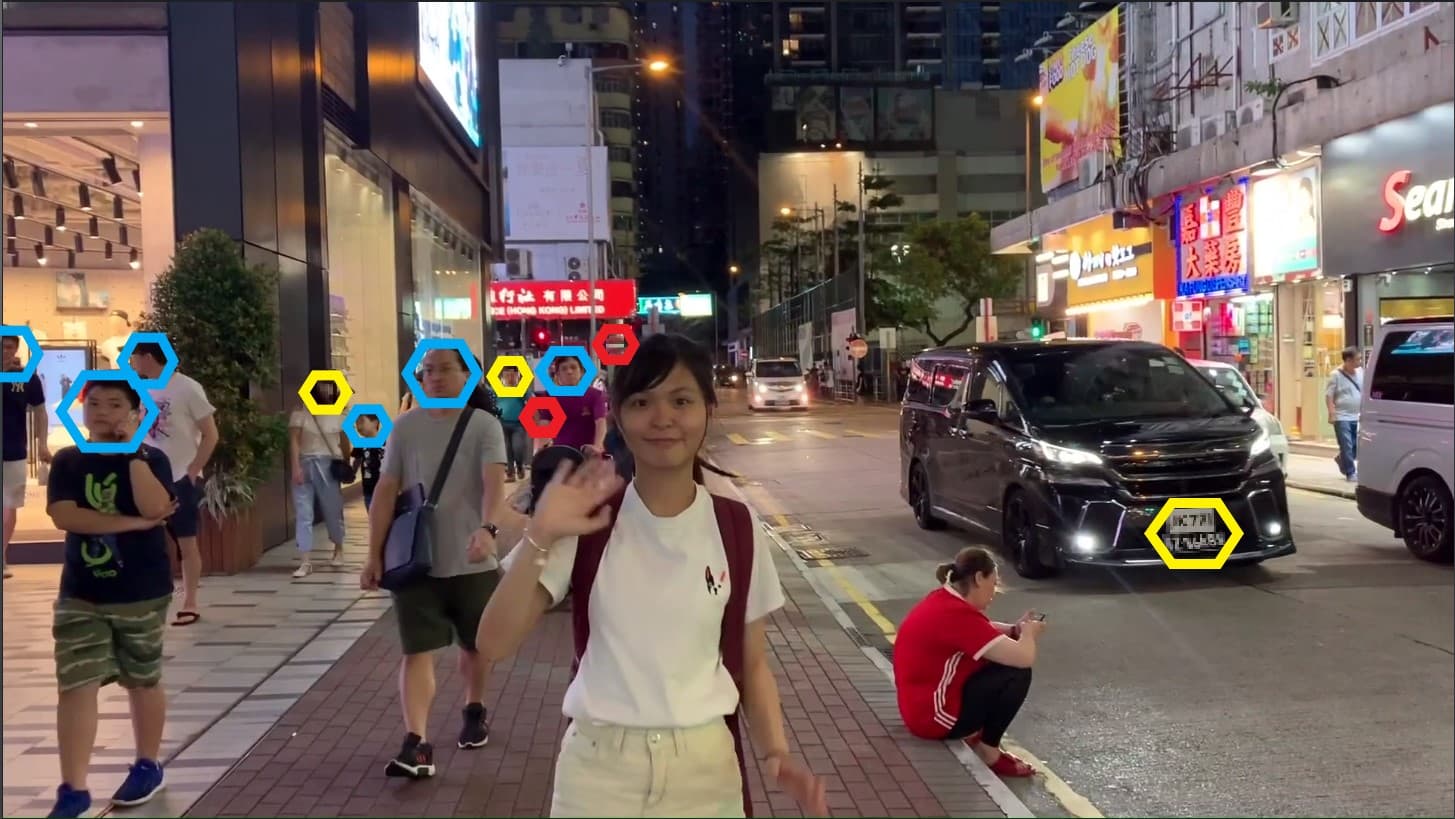}\\
\vspace{-0.2em}
\includegraphics[width=28.4mm]{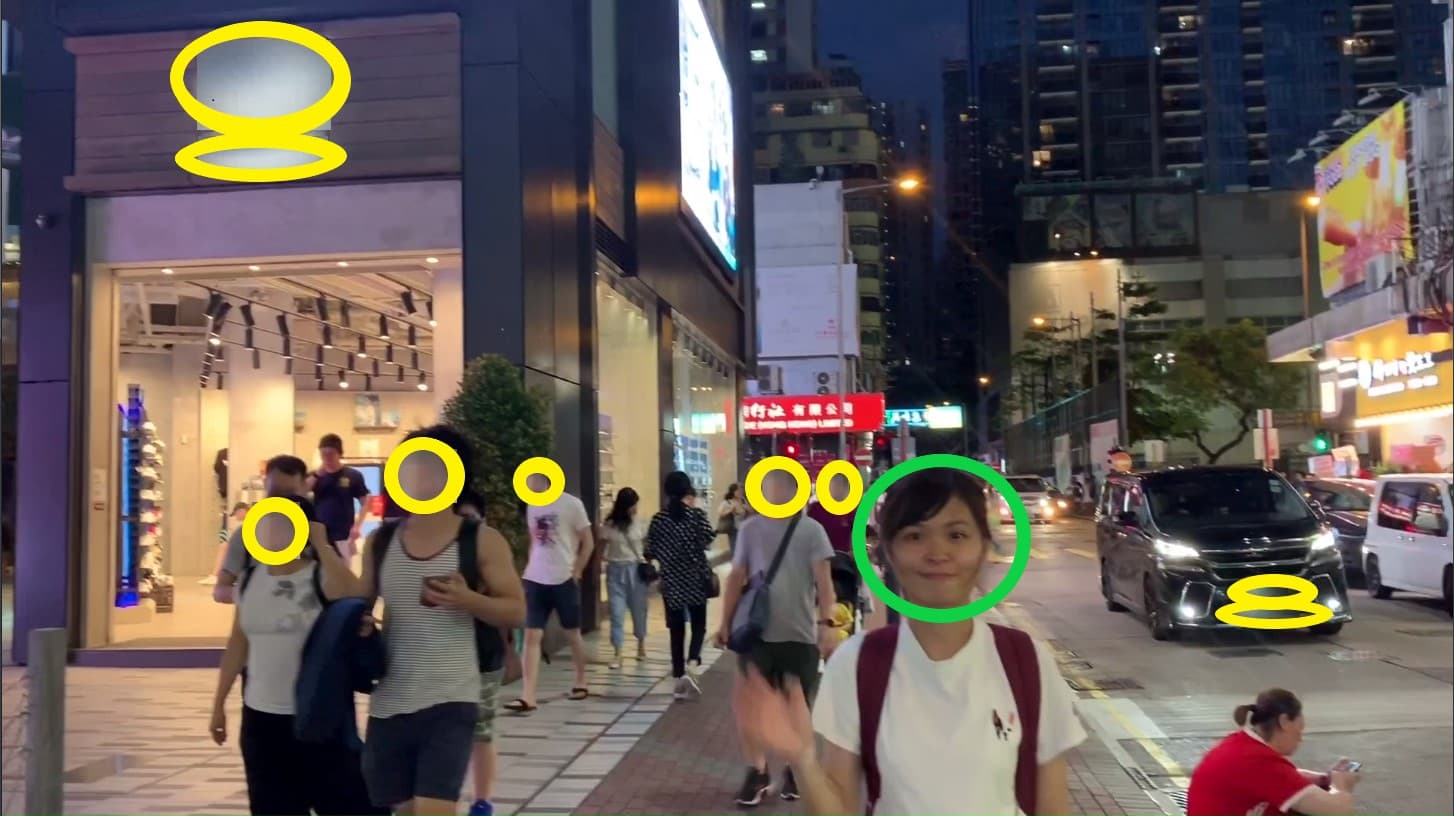}& \hspace{-0.8em}
\includegraphics[width=28.4mm]{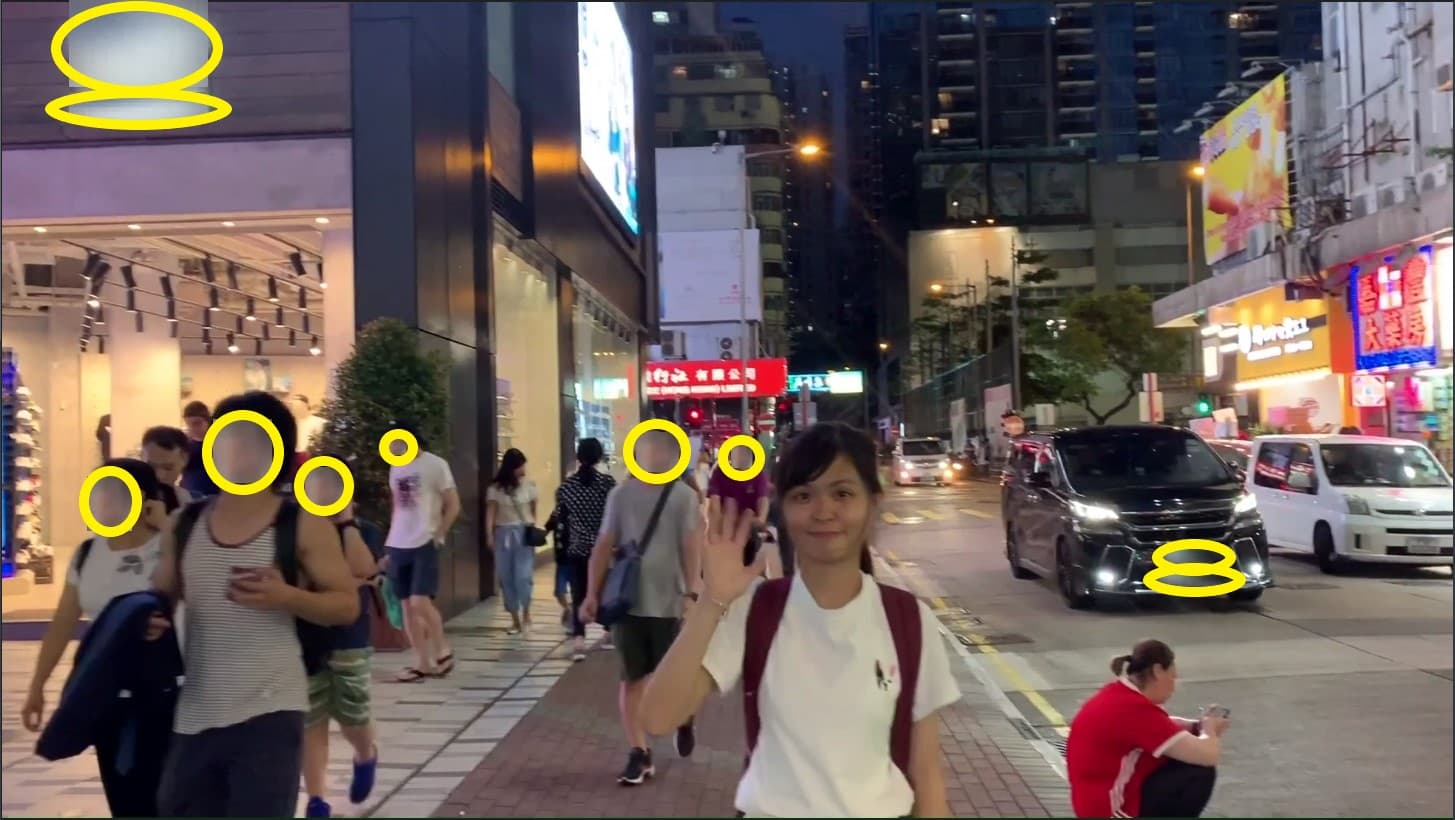}& \hspace{-0.8em}
\includegraphics[width=28.4mm]{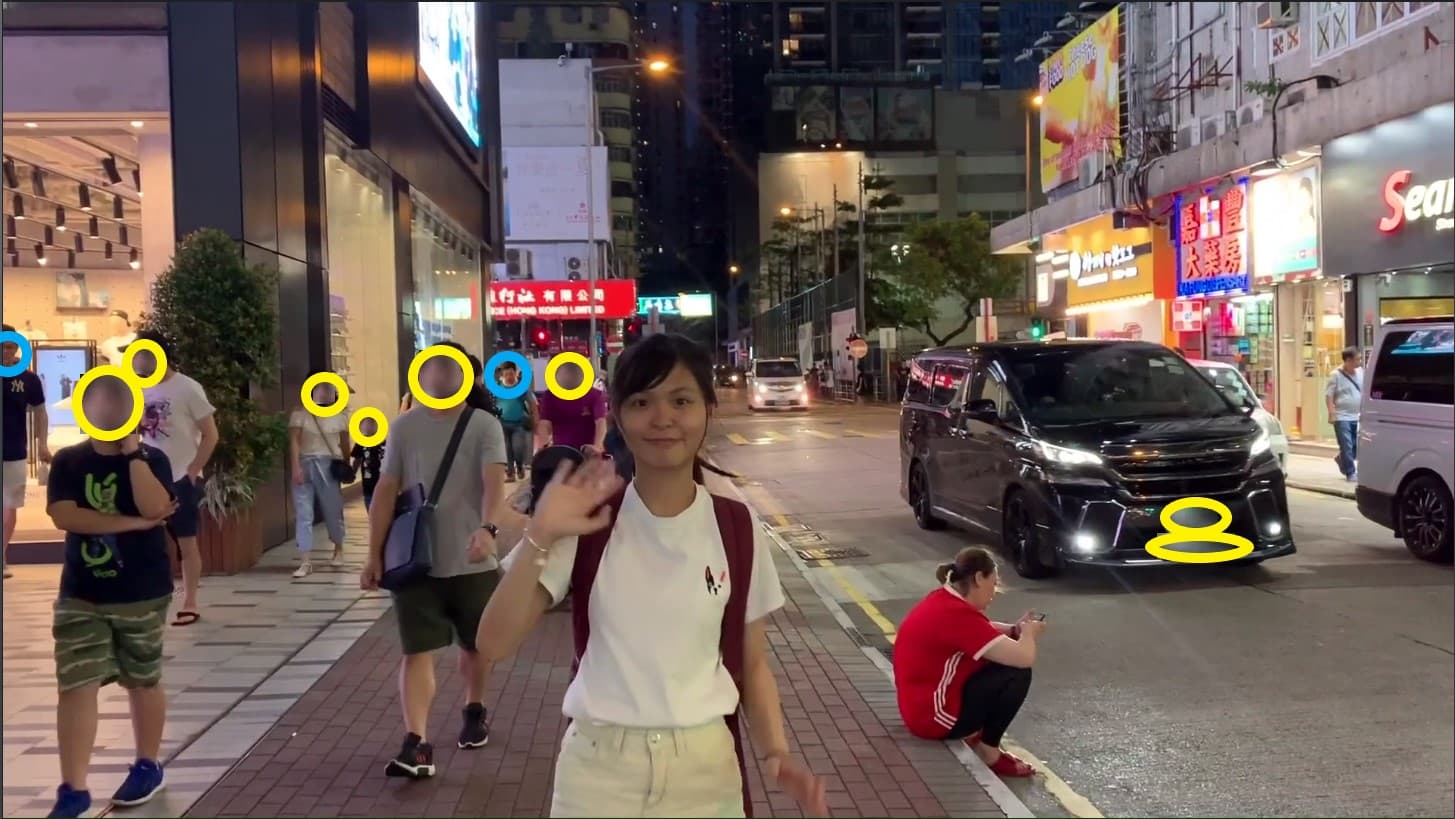}
\end{array}$
\end{center}
\vspace{-0.5em}
\caption{Qualitative comparison of YouTube (mosaics in rectangular), KCF (mosaics in hexagons), and PsOP (mosaics in circles). Green boxes indicate human intervention is needed for specification, yellow are the auto-generated mosaics, red ones are the over-pixelated area, and blue ones are the miss-pixelated area.}
%\label{pics:PIAP example}
\vspace{-1em}
\end{figure}

\par

Table 4 is the face clustering performance of classic AP, PAP (AP after position revision), and PIAP on the collected live streaming dataset. Comparing AP with PAP, the position information of face vectors significantly boosts the purity of the clustering result. According to the results of PIAP and AP, our incremental message passing algorithm solves the time-consuming problem of AP almost without affecting the purity as we stated before. The value of the Time column is the seconds needed to reach an iteration of consensus.

\subsection{Qualitative Comparison}
Qualitative comparisons among the YouTube offline pixelation tool, migrated KCF, and PsOP are shown in Figure 2. Snapshots are shoot from left to right. IPOs, including the trademark and the car plate number, and DPOs, including faces and texts, are contained in the scenes. The first row is the original scene in which a girl streamer on the street is waving hands to greeting the audience. The 2nd, 3rd, and 4th rows are respectively the pixelation results after YouTube, KCF, and PsOP. Besides the significant differences in red and blue boxes indicating accuracy, the green boxes in the leftmost snapshot demonstrate the huge gap of required manual work on the initialization. Only in PsoP, privacy-sensitive objects are auto-detected and categorized, and DPOs are clustered and pixelated with minimum intervention.
\iffalse
\subsubsection{Efficiency}
Training and testing are conducted on our i7-7800X, 64G RAM and dual GTX 1080 machine. The main cost of PsOP is on the face embedding algorithm. The embedding for one face takes 10-15ms. So under extreme circumstances that contain many faces, we can trick the detection and embedding jumping among frames for efficiency.
\fi
\subsection{Efficiency}
Training and testing are conducted on our i7-7800X, 64G RAM and dual Nvidia GTX 1080 machine. The main cost of FPVLS is on the face embedding algorithm. The embedding for one face takes 10-15ms on our i7-7800X, dual GTX1080, 32G RAM machine. The face detection, including compensation for a frame, takes 3ms. Another time-costing part is the initialization of PIAP. The initialization process takes 10-30ms depending on the case. Furthermore, each incremental propagation loop takes 3-5ms. So PsOP can satisfy the real-time efficiency requirement. In the cases of extreme circumstances that the live video streams contain many faces or other privacy sensitive objects, we can reduce the sampling rate of the video frames to improve the efficiency of the detection and recognition processes.

\section{Conclusion}
In this paper, a novel extendable and unified framework using Privacy-sensitive Objects Pixelation (PsOP) is proposed for live video streaming. The proposed PsOP shows significant improvements in pixelation accuracy, precision, and over-pixelation ratio compared with other offline pixelation algorithms. Also, our PsOP requires minimum human intervention on the pixelation of discriminating pixelation objects represented by faces and texts. Leveraging the trained deep CNNs, the proposed PsOP manages to solve the pixelation problems in video live streamings through the proposed Positioned Incremental Affinity Propagation (PIAP) clustering. The PIAP deals with the ill-defined cluster number issue and boosts the accuracy of classic Affinity Propagation (AP) through position information. Moreover, the accuracy of AP is not affected by the incremental processing.  However, the low-resolution video streamings are still quite challenging because the image resolution affects much of the detection and embedding network performance. Designed and trained on high-quality stationary images, these deep neural networks can drop to a deficient performance and break the robustness of the PsOP. Our future work will focus on the improvement of accuracy and robustness under low resolution streaming scenes.

\begin{acks}
This work was partly supported by the University of Macau under Grants: MYRG2018-00035-FST and MYRG2019-00086-FST, and the Science and Technology Development Fund, Macau SAR (File no. 041/2017/A1, 0019/2019/A).
\end{acks}

\bibliographystyle{ACM-Reference-Format}
\bibliography{sample-base}

\appendix
\iffalse
\section{Research Methods}

\subsection{Part One}

Lorem ipsum dolor sit amet, consectetur adipiscing elit. Morbi
malesuada, quam in pulvinar varius, metus nunc fermentum urna, id
sollicitudin purus odio sit amet enim. Aliquam ullamcorper eu ipsum
vel mollis. Curabitur quis dictum nisl. Phasellus vel semper risus, et
lacinia dolor. Integer ultricies commodo sem nec semper.

\subsection{Part Two}

Etiam commodo feugiat nisl pulvinar pellentesque. Etiam auctor sodales
ligula, non varius nibh pulvinar semper. Suspendisse nec lectus non
ipsum convallis congue hendrerit vitae sapien. Donec at laoreet
eros. Vivamus non purus placerat, scelerisque diam eu, cursus
ante. Etiam aliquam tortor auctor efficitur mattis.

\section{Online Resources}

Nam id fermentum dui. Suspendisse sagittis tortor a nulla mollis, in
pulvinar ex pretium. Sed interdum orci quis metus euismod, et sagittis
enim maximus. Vestibulum gravida massa ut felis suscipit
congue. Quisque mattis elit a risus ultrices commodo venenatis eget
dui. Etiam sagittis eleifend elementum.

Nam interdum magna at lectus dignissim, ac dignissim lorem
rhoncus. Maecenas eu arcu ac neque placerat aliquam. Nunc pulvinar
massa et mattis lacinia.
\fi
\end{document}